\documentclass[10pt, preprint, 3p]{elsarticle}
\usepackage{graphicx}
\graphicspath{{./fig/}}
\DeclareGraphicsExtensions{.pdf,.jpeg,.png}
\usepackage[T1]{fontenc}
\usepackage{mathtools}
\usepackage{amsmath}
\usepackage{amssymb}
\usepackage{amsthm}

\usepackage{rotating}
\usepackage{multirow}
\usepackage{float} %%% pma: wichtig!!!
\usepackage{balance}
\usepackage{tabularx}
\usepackage{hyperref}
\usepackage{bm}
\usepackage{soul}
\usepackage[dvipsnames]{xcolor}
\usepackage{colortbl}
\soulregister\cite7
\soulregister\citeRest7
\soulregister\ref7
\soulregister\pageref7
\setstcolor{red}
\usepackage{algorithm}
\usepackage{adjustbox}

\usepackage{algpseudocode}
\usepackage{siunitx}
\usepackage{mathtools}
\usepackage{enumitem}
\usepackage{tikz}
\usetikzlibrary{decorations.markings}
\usetikzlibrary{backgrounds}
\newcommand{\boundellipse}[3]% center, xdim, ydim
{(#1) ellipse (#2 and #3)
}
\definecolor{jblue}{rgb}{0.008, 0.788, 0.978}

\tikzstyle{decision} = [diamond, draw, text badly centered, inner sep=3pt]
\usetikzlibrary{shapes.geometric}

\definecolor{myG}{rgb}{0.13, 0.55, 0.13}
\definecolor{myB}{rgb}{0.0, 0.45, 0.73}
\definecolor{myR}{rgb}{0.82, 0.1, 0.26}
\usepackage[column=Q]{cellspace}
\usepackage{adjustbox}
\usepackage{array}
\usepackage{booktabs}
\usepackage{multirow}
\usepackage{subcaption}

\newcolumntype{R}[2]{%
	>{\adjustbox{angle=#1,lap=\width-(#2)}\bgroup}%
	l%
	<{\egroup}%
}
\newcommand{\nonl}{\renewcommand{\nl}{\let\nl\oldnl}}
\newcommand\mdoubleplus{\mathbin{+\mkern-10mu+}}
\title{Temporal Convolution Derived Multi-Layered Reservoir Computing}
\author[1]{Johannes Viehweg\corref{cor1}\fnref{fn1}}
\ead{johannes.viehweg@tu-ilmenau.de}
\author[1]{Dominik Walther\fnref{fn1}}
\ead{dominik.walther@tu-ilmenau.de>}
\author[1,2]{Patrick M\"ader\fnref{fn1,fn2}}
\ead{patrick.maeder@tu-ilmenau.de}

\cortext[cor1]{Corresponding author: johannes.viehweg@tu-ilmenau.de}

\address[1]{Helmholtzplatz 5, 98693 Ilmenau, Germany}
\address[2]{Philosophenweg 16, 07743 Jena, Germany}

\fntext[fn1]{Johannes Viehweg, Dominik Walther, and Patrick Mäder are with the Data-intensive Systems and Visualization group (dAI.SY), Technische Universit\"at Ilmenau, 98693 Ilmenau, Germany.}
\fntext[fn2]{Patrick Mäder is with the Faculty of Biological Sciences, Friedrich Schiller University, 07743 Jena, Germany.}

\begin{document}
	\begin{abstract}
		
		The prediction of time series is a challenging task relevant in such diverse applications as analyzing financial data, forecasting flow dynamics or understanding biological processes. Especially chaotic time series that depend on a long history pose an exceptionally difficult problem. While machine learning has shown to be a promising approach for predicting such time series, it either demands long training time and much training data when using deep Recurrent Neural Networks. Alternative, when using a Reservoir Computing approach it comes with high uncertainty and typically a high number of random initializations and extensive hyper-parameter tuning. In this paper, we focus on the Reservoir Computing approach and propose a new mapping of input data into the reservoir's state space. Furthermore, we incorporate this method in two novel network architectures increasing parallelizability, depth and predictive capabilities of the neural network while reducing the dependence on randomness. For the evaluation, we approximate a set of time series from the Mackey-Glass equation, inhabiting non-chaotic as well as chaotic behavior as well as the SantaFe Laser dataset and compare our approaches in regard to their predictive capabilities to Echo State Networks, Autoencoder connected Echo State Networks and Gated Recurrent Units. For the chaotic time series, we observe an error reduction of up to $85.45\%$ compared to Echo State Networks and $90.72\%$ compared to Gated Recurrent Units. Furthermore, we also observe tremendous improvements for non-chaotic time series of up to $99.99\%$ in contrast to the existing approaches.
		
	\end{abstract}
	\maketitle
	
	% !TEX root = ../main.tex
	
	\section{Introduction}
	\label{sec:intro}
	Time series prediction plays a crucial role in various fields, encompassing biologically inspired systems such as population dynamics \cite{kuang1993delay} and financial stock market metrics \cite{makridakis2020m4} up to climate and global flow models \cite{zhai2024machine, panahi2024machine}. However, predicting time series is a challenging task, particularly when dealing with chaotic, non-periodic dynamics. 
	In the pursuit of accurate predictions, machine learning techniques, particularly Recurrent Neural Networks (RNNs) such as Long-Short Term Memory Networks (LSTM) \cite{hochreiter1997long} and the subsequent Gated Recurrent Units (GRU) \cite{cho2014properties}, have emerged as a viable approach for predicting dynamical systems \cite{teutsch2022flipped} but suffer from their demand of time and resources for the training \cite{vlachas2020backpropagation}. 
	Feed-forward-based Transformer architectures \cite{vaswani2017attention} allowed to overcome these problems, e.g., by parallelizing the training process, especially in the area of natural language processing. Another approach to be named here is the idea of convolutions in time, namely Temporal Convolutional Networks (TCN) \cite{bai2018empirical}.
	
	However, they have not been found superior when predicting chaotic time series,  e.g., shown in \cite{qin2019comparison, shahi2022prediction, fu2023ibm}.
	
	In recent years, Reservoir Computing (RC) \cite{verstraeten2007experimental}, especially Echo State Networks (ESN) \cite{jaeger2001echo} with only three layers, namely the input layer, a single recurrent reservoir and the output layer, have garnered significant attention due to their effectiveness in time series prediction and their low resource demand in training and prediction phase \cite{pathak2018hybrid, lu2017reservoir}. A problem of the approach is the high variance between training results due to the weights being chosen randomly and, except for the ones of the output layer, not being further optimized. This can lead to a heterogenous situation with some trained networks being well fitted and others not resembling the target in a sufficient manner while being trained with the same hyper-parameters but different drawn weights. One approach of handling this problem is the adaptation of the structure of the reservoir to a specific problem \cite{sun2022systematic} or utilizing additional methods of regularization \cite{zhai2023emergence}.
	Several previous studies propose the concept of multiple reservoir layers rather than using a single one in order to improve ESN's prediction performance, e.g. \cite{malik2016multilayered, moon2021hierarchical}. However, we argue that despite improved prediction performance for luckily drawn configuration, this approach further increases the randomness due to multiple randomly chosen reservoirs. 
	Other recent works in the field of RC deliver a new theoretical foundation and explanation of the approach \cite{bollt2020explaining} and argue that their effectiveness could be explained by showing their connection to Vector Autoregressive Average (VAR) for the identity activation function. This allows for a predictor without the need to rely on favorable randomness for the weights. It also allows for an analysis of the ESN, which would be unfeasible for the basic case. 
	According to \cite{gauthier2021next}, an optimized RC network is hereby equivalent to non-linear VAR, which allows for construction of a predictor without randomness. This understanding is used to propose novel architectures such as the Next Generation Reservoir Computing (NG-RC) \cite{gauthier2021next} reducing the needed randomness. 
	In conclusion, previous studies observe instabilities for long delays, i.e., mapping a dependence of a long time interval of the time series into the network's reservoir, and a dependence on randomness in the initialization of the network.
	
	We adapt the theoretical foundation of reservoir computing discussed above and propose a new architecture and mapping to overcome the discussed problems with randomness and long delays both encouraging an unstable autoregressive prediction resulting in a constant predicted value or an non-computable error.
	%, stable even for long inputs, paired with a multi-layered approach on such previously hard tasks. 
	We show a first stable auto-regressive prediction, staying close to the ground truth, for multiple Lyapunov times for those time series while exploiting the benefits of parallelizability offered by its inspiration, the TCNs and further increasing the predictive capabilities by coupling it with an Extreme Learning Machine (ELM) of \cite{huang2004extreme} into a state space which we argue benefits from being in the field of so called hyper-dimensional computing \cite{kanerva2009hyperdimensional} similar to \cite{gallicchio2011architectural}. 
	%It is also the first in proposing this additional mapping into a high dimensional state space with an selection of its size, potentially a solution to simplify the tuning of the size of the reservoir.
	\\
	Our research aims to answer the following questions:
	\begin{enumerate}
		\item How can the problem of randomness be mitigated without losing the ability to handle time series with an importance of memory?
		\item Is the adaptation of the idea of Temporal Convolutional Networks suitable for the handling of chaotic time series?
		\item Is the introduction of randomness beneficial in the case of chaos?
	\end{enumerate}
	
	To answer those we evaluate our approach on multiple configurations for well-established benchmark datasets. Those are chosen for assessing the predictive capabilities of algorithms in chaotic, history-dependent systems. We generated them based on the the Mackey-Glass equation \cite{mackey1977oscillation, glass2010mackey}, which has been widely used in studies such as \cite{bianchi2016investigating, yao2020fractional, li2020deep, li2018echo, morales2021unveiling} as well as utilizing the SantaFe Laser dataset \cite{weigend1993results}.
	This paper is structured as follows:
	In Section~\ref{sec:RelW} we discuss the different baseline varieties of RC and ESN used in this work. Our proposal and background to it are explained in Section~\ref{sec:TC_chapter}. The baselinea of AEESN and GRUs as well as the data is shown in Section~\ref{sec:eval} together with the evaluation and results of our tests. We close the paper with a discussion of its limitation in Section~\ref{sec:discussion} and conclude in Section~\ref{sec:concl}.

	% !TEX root = ../main.tex

	\section{Reservoir Computing}\label{sec:RelW}
	The core principle of RC involves reducing the complexity of training by focusing on a limited set of weights. These weights are crucial for linearly mapping a high-dimensional state to a specific output in a singular computation step. Within this framework, an autoregressive model for time series prediction is established over a test interval of length $S_P$, following a training and initialization phase of lengths $S_T$ and $S_I$, respectively. This model handles inputs $x$ and targeted outputs $y$, for the single step prediction as $x^{(t'+1)}=y^{(t')}, t'\geq S_I+S_T$. In case of the multi step prediction and an computed output $\hat{y}$ we assume from this point forward the equation  $x^{(t'+1)}=\hat{y}^{(t')}$.

	\begin{figure}
		\centering
		%	\fbox{
		\includegraphics[trim=110 70 70 40,clip, height=.25\textheight]{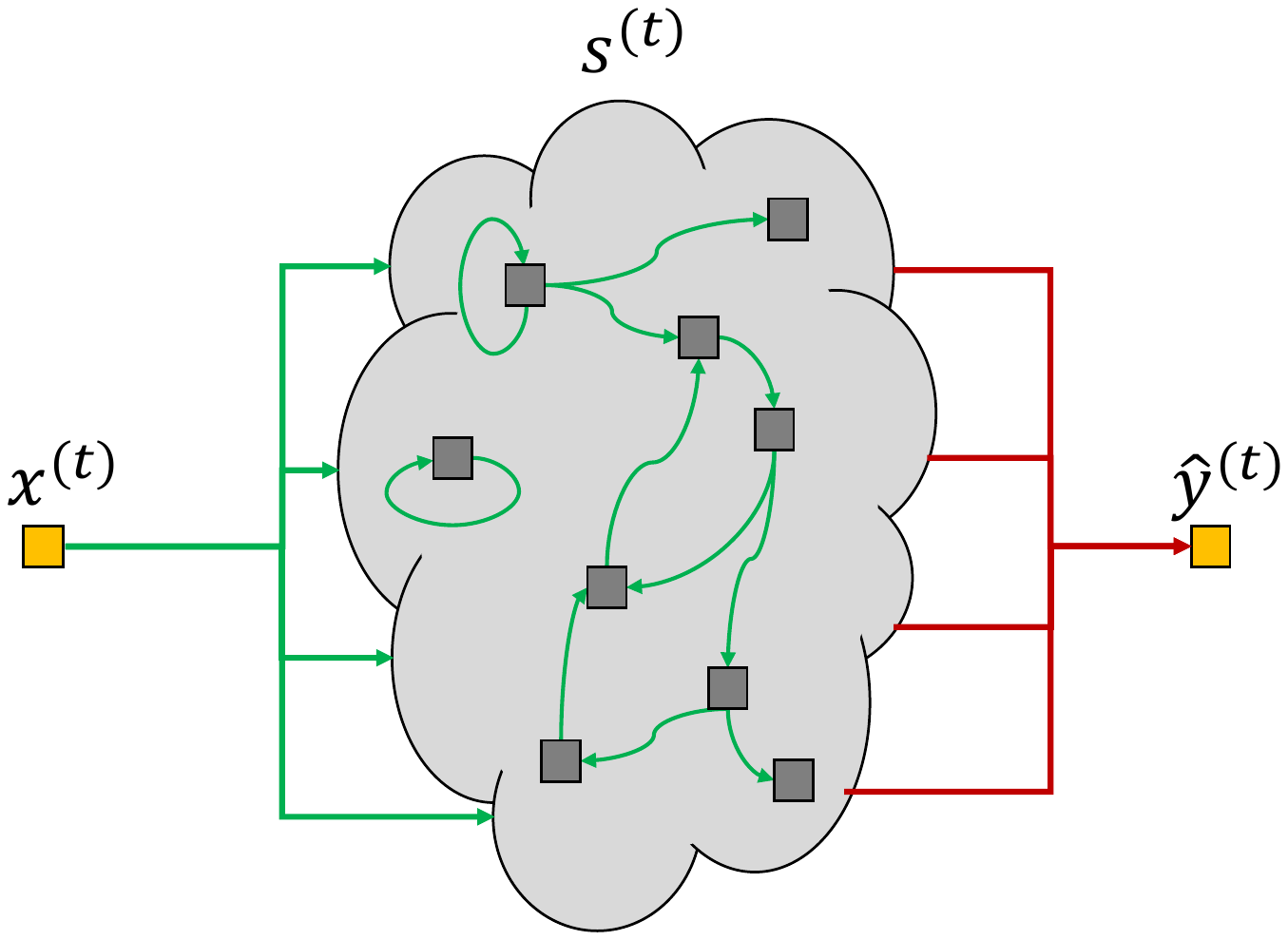}
		%	}
		\caption{Basic architecture of the ESN, where green arrows ({\color{OliveGreen}$\rightarrow$}) refer to the randomly initialized set of connections from the input $x^{(t)}$ into the reservoir and red arrows ({\color{red}$\rightarrow$}) refer to the trained mapping from the reservoir to the output $\hat{y}^{(t)}$}.
		\label{fig:ESN}
	\end{figure}
	
	\subsection{Echo State Networks}\label{sec:ESN}
	In this paper, our focus is on the extensively studied ESN architecture as most well known and used example of RC. In its basic form, depicted in Figure~\ref{fig:ESN}, the ESN consists of three layers with respective weights $W^{\mathrm{in}}\in\mathbb{R}^{N^{\mathrm{r}} \times N^{\mathrm{in}} }$, $W^{\mathrm{r}}\in \mathbb{R}^{N^{\mathrm{r}}\times N^{\mathrm{r}}}$ and $W^{\mathrm{out}}\in\mathbb{R}^{N^{\mathrm{out}}\times N^{\mathrm{r}}}$. Here, $N^{\mathrm{in}} = \hat{\delta}N^{\mathrm{out}}$ with $\hat{\delta}$ denoting an input delay, typically chosen as $\hat{\delta}=1$, and $N^{\mathrm{in}}<<N^{\mathrm{r}}$. The weights $W^{\mathrm{in}}$ and $W^{\mathrm{r}}$ are randomly generated but fixed, as suggested in \cite{jaeger2001echo, lukovsevivcius2009reservoir, lukovsevivcius2012practical, jaeger2007echo}. The weights are drawn from a uniform distribution: $\forall w \in W^{[\mathrm{in}, \mathrm{r}]}, w \sim \mathcal{U}(-\delta, \delta)$, with $\delta\in\mathbb{R}$ and a connectivity between the positions in the state space $\sigma(W^{\mathrm{r}})\in\mathbb{R}_{w^r\in[0,1]}$. Only the output mapping using the weights $W^{\mathrm{out}}$ is learned. The recurrent connections enable memory in the network, typically scaled by the number of neurons $N^{\mathrm{r}}\in\mathbb{N}$ and the spectral radius $\rho\in\mathbb{R}{\geq0}$ of the recurrent weights $W^{\mathrm{r}}$. The internal weights in the matrix $W^{\mathrm{r}}$ are scaled by the spectral radius $\rho$, often chosen to be slightly below $1$ \cite{jaeger2007optimization,lukovsevivcius2012practical}. For the RC approaches $x^{(t)}\in\mathbb{R}^{N^{\mathrm{in}}}$ is the input at time step $t\in\mathbb{N}{>0}$, $s^{(t)}\in\mathbb{R}^{N^{\mathrm{r}}}$ the state, and $\hat{y}^{(t)}\in\mathbb{R}^{N^{\mathrm{out}}}$ the output as shown in Fig.~\ref{fig:ESN}. The state $s^{(t)}$ is calculated as follows:
	\begin{equation}
	s^{(t)} = f(W^{\mathrm{in}}x^{(t)}+W^{\mathrm{r}}s^{(t-1)}),
	\end{equation}
	where $f(\cdot)$ represents a non-linear activation function $f(\cdot) = \tanh(\cdot)$.
	
	The output weights $W^{\mathrm{out}}$ are the only weights that are learned. This is accomplished in a single step using ridge regression, also known as Tikhonov regularization, as proposed in \cite{jaeger2001echo}. The initial state is used to compute the first $S_I$ steps, which are then discarded during the learning process. Subsequently, the targets $Y\in \mathbb{R}^{N^{\mathrm{out}}\times S_T}$ and states $S\in\mathbb{R}^{N^{\mathrm{r}}\times S_T}$ for the next $S_T$ time steps are collected. The computation of $W^{\mathrm{out}}$ is done as:
	\begin{equation}\label{eq:Wout}
	W^{\mathrm{out}} = YS^T(SS^T+\beta \mathbb{I})^{\dagger},
	\end{equation}
	where $\beta$ is the regularization coefficient, $\dagger$ denotes the Moore-Penrose pseudoinverse, and $\mathbb{I}$ represents the identity matrix of size $N^{\mathrm{r}}\times N^{\mathrm{r}}$. Alternative approaches to learning $W^{\mathrm{out}}$, such as the Least Absolute Selection and Shrinkage Operator (LASSO) \cite{tibshirani1996regression} or Support Vector Machines (SVM) \cite{boser1992training, cortes1995support}, are also viable options. Other methods of regularization are discussed in the literature, such as the introduction of noise \cite{lukovsevivcius2009reservoir} but are rarely observed to be used in application studies \cite{VIEHWEG2022}. Therefore, we limit the regularization of the ESN to the use of $\beta$ in this work. The random initialization and computation of states introduce variance in the values of the weights and predictions. Although hyper-parameter optimization can reduce the variance between initializations \cite{VIEHWEG2022}, multiple runs per task are necessary. To optimize the networks for a specific task a set of hyper-parameters must be chosen $\Omega = \{N^{\mathrm{r}}, \rho, \gamma, \beta, S_T, S_I\}$. In the frame of this work we refer to this approach as Basic ESN.
	
	\subsection{Extreme Learning Machines}\label{sec:ELM}
	A method related to RC, but without recurrence and holding a network's state, aka memory, are Extreme Learning Machines (ELM) \cite{huang2004extreme}. An analogous method of learning the output layer's weights $W^{\mathrm{out}}$ is used. In contrast to the Basic ESN the state is computed as: 
	\begin{equation}
	s^{(t)} = f(W^{\mathrm{in}}x^{(t)}).
	\end{equation}
	
	Still multiple runs with different random seeds are necessary to determine the error of the prediction with certainty. Approaches of multiple layers, while used in other neural networks \cite{teo2001wavelet}, are still an active area of research for RC \cite{malik2016multilayered, ma2020deepr}.

	\subsection{Next Generation Reservoir Computing}\label{sec:NGRC}
	An obvious problem of the prior discussed networks is their initialization based on random distributions. This typically results in significant performance variations between differently initialized models, subsequently implying the need for multiple training runs to determine the most suitable model and its predictive capabilities. Several approaches have been proposed to consider this randomness and to offer more predictable training results, most notable Gauthier et al.'s Next Generation Reservoir Computing (NG-RC) \cite{gauthier2021next}. They substitute the randomly initialized mapping into the state space by a deterministic non-linear function defined e.g. as: 
	\begin{equation}
	s^{(t)} = \hat{f}(g(\mdoubleplus_{i=0}^{\hat{\delta}} x^{(t-i)})).
	\end{equation}
	
	The authors propose e.g. a concatenation of $\hat{\delta}$ elements to a vector $s^{(t)} \in\mathbb{R}^{N^{\mathrm{r}}\times 1}$ as $\mdoubleplus_{i=0}^{j}$. Here, $g(\cdot)$ refers to the calculation on the inputs and $\hat{f}(\cdot)$ is the activation function, chosen as identity function. 
	Analogous to the ESN, the states $s^{(k)}$ for $0\leq k<S_T$ are saved in a matrix for all $S_T$ time steps of training, but due to the feed-forward nature without the need to discard an initial wash-out phase of $S_I$ steps. The training of the output weights remains analogous to the standard ESN training (cp. Eq.~\ref{eq:Wout}). The novel idea proposed by NG-RC is a mapping constructed from a linear part $\mathbb{O}_{lin}^{(\cdot)}$ and a non-linear one $\mathbb{O}_{nonlin}^{(\cdot)}$. The linear part is a concatenation of all delayed inputs as 
	\begin{equation}
	\mathbb{O}_{lin}^{(t)} = \mdoubleplus_{i=0}^{\hat{\delta}}x^{(t-i)}.
	\end{equation}
	
	For the non-linear part, the upper right triangle matrix of the outer product is used, which is given as $(\cdot) \otimes \rceil (\cdot)$ and concatenated with the squared values of the linear part as
	\begin{equation}
	\mathbb{O}_{nonlin}^{(t)} = [\mdoubleplus_{i=0}^{\hat{\delta}}x^{{(t-i)}^2};\mathbb{O}_{lin}^{(t)} \otimes \rceil \mathbb{O}_{lin}^{(t)}]. 
	\end{equation}
	
	For $g(\cdot)$ a mapping is assumed in the form
	\begin{equation}
	g(\mdoubleplus_{i=0}^{\hat{\delta}} x^{(t-i)}) = [\mathbb{O}_{lin};\mathbb{O}_{nonlin}],
	\end{equation}
	with $[\cdot;\cdot]$ being the concatenation of the given inputs.
	
	\section{Adapting Temporal Convolutions for Reservoir Computing}
	\label{sec:TC_chapter}
	Temporal convolutions emerge as a significant advancement within feed-forward networks, reignited by the successes in Convolutional Neural Networks (CNNs) for image processing \cite{lecun1989generalization} and Transformers for language processSing tasks \cite{vaswani2017attention}. Historically, time series analysis has been dominated by Recurrent Neural Networks (RNNs). However, recent empirical studies \cite{bai2018empirical} have demonstrated the efficacy of Temporal Convolutional Networks (TCNs) in this domain, particularly in tasks such as classification \cite{walther2022automatic}, where they have achieved notable success. TCNs operate by processing inputs within a defined window size in a pairwise fashion to generate outputs, with the stride between inputs incrementally increasing as the network deepens. Importantly, the process of optimizing the weights assigned to each input for subsequent layers is conducted through backpropagation, ensuring effective learning and adaptation to time series data.
	
	\subsection{The Temporal Convolution Concept}
	\label{sec:TC}
\begin{figure}[htb]
		\centering
		\includegraphics[trim=0 0 0 0, clip,height=.3\textheight]{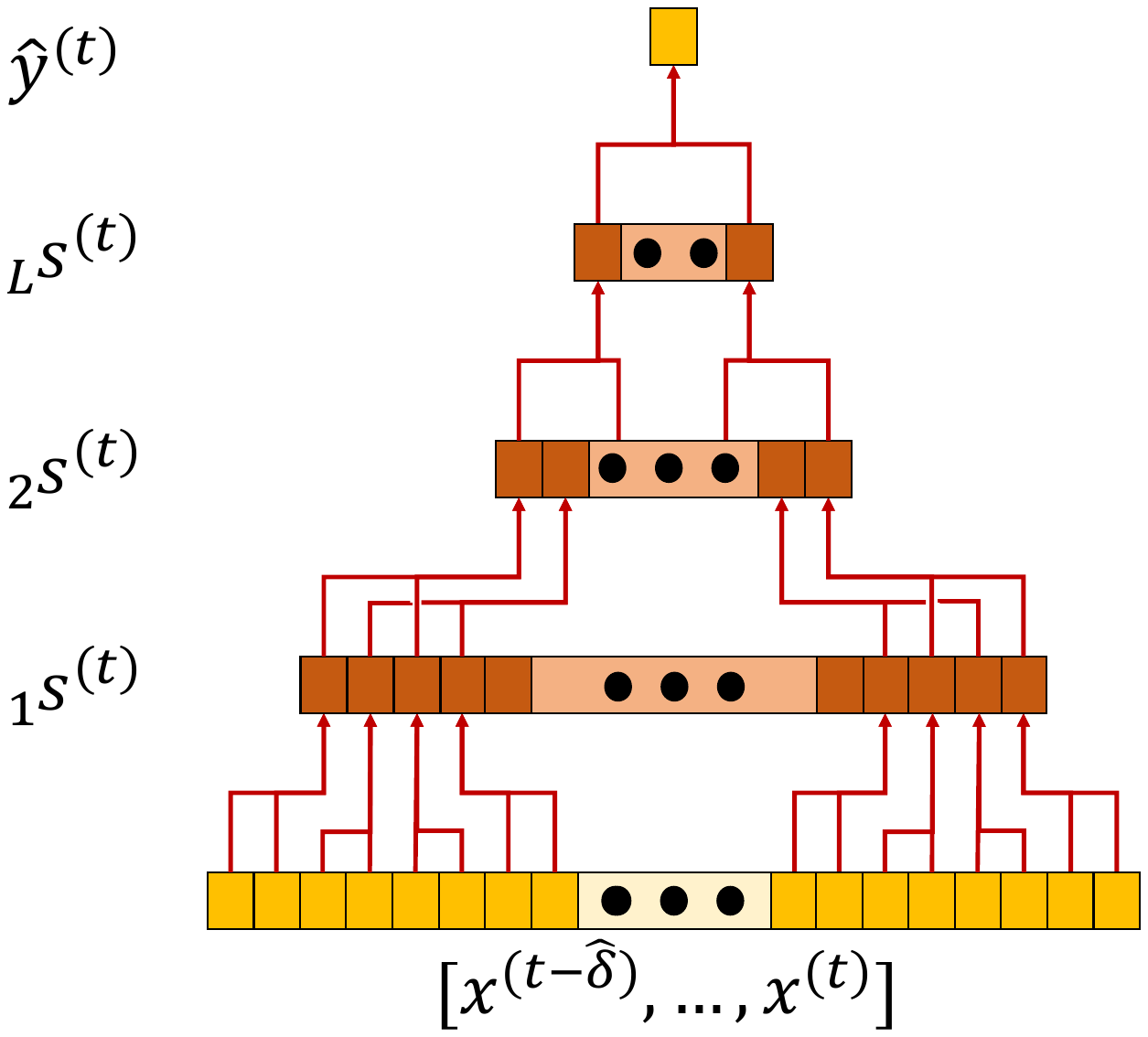}
		\caption{Single block of a TCN $\chi_i$, where red arrow-style edges ({\color{red}$\rightarrow$}) refer to the information flow from the input through multiple layers of convolutions to each subsequent layer as well as from the last layer $L$ to the output $\hat{y}^{(t)}$.}
		\label{fig:TCN}
	\end{figure}

	Temporal Convolutions \cite{walther17systematic} combine the benefits of a feed-forward network, i.e., parallelizable computation and efficient training, with the benefits of recurrent networks, i.e., maintaining a history of previous networks states facilitating the analysis of long-term dependencies in the input data. The inputs $x^{(t)}, x^{(t-1)}$ are weighted together in the first layer, which results in one value in the state space. We refer to this as a token $\tilde{x}_k\in {}_ls_k^{(t)}$ in this work, i.e. the $k$-th element of the $l$-th layer. In the succeeding layer, two such computed tokens $\tilde{x}_{(k)}$ and $\tilde{x}_{(k-2)}$ are used, skipping $\tilde{x}_{(k-1)}$. For each layer $l\geq 1$ this implies a computation of tokens from the former layer, generalized as $\hat{x}= {}_{l-1}s^{(t)}$, in form of the use of $\hat{x}_{(k)}$ and $\hat{x}_{(k-2^{l-1})}$, respectively with $l=0$ being the original input. Figure~\ref{fig:TCN} visualizes that mechanism and shows that convolutions only have access to current and previous time steps. Hereby, Figure~\ref{fig:TCN} shows a single block $\chi_i$ whereas a TCN is build from $m$ such block, $1\leq i\leq m$. In this direct mapping the history of the time series is free of its sequential order for $\hat{\delta}$ time steps. This allows for a parallelization of the computation, resulting in TCNs being able to exploit the benefits of feed-forward networks.

	\subsection{Temporal Convolution Derived Reservoir Computing (TCRC)}\label{sec:TCRC}
	The basic idea of the approach is to map the input $x^{(\cdot)}$ into the state $s^{(\cdot)}$ by a pairwise multiplication of $x^{(t-i)}\cdot x^{(t-i-1)}$. The similarity to NG-RC is easy to see, but the multiplication scheme simplified, the resulting state space subsequently reduced. In this paper we will refer to this as Temporal Convolution Derived Reservoir Computing (TCRC). The idea is comparable albeit more limited in the resulting connections to the NG-RC approach (cp. Sec.~\ref{sec:NGRC}) by accessing only inputs of direct neighborhoods rather than all inputs in the range of the delay. We assume for the first network layer 
	\begin{equation}
	{}_1s^{(t)}_i = x^{(t-i)}x^{(t-i-1)}.
	\end{equation}
	Furthermore, we assume  $0\leq i < \hat{\delta}<S_T$, with $\hat{\delta}$ being the maximum delay applied to the input. The state vector of the first layer ${}_1s^{(t)}$ is calculated as: 
	\begin{equation}\label{eq:first_TC_state}
	{}_1s^{(t)} = \mdoubleplus_{i=1}^{\hat{\delta}}{}_1s_{i}^{(t)}%[\mdoubleplus_{i=1}^{\hat{\delta}}x^{(t-1)};\mdoubleplus_{i=1}^{\hat{\delta}}{}_1s_{i}^{(t)}].
	\end{equation}
	Here, $\mdoubleplus_{i=1}^{k}$ is the concatenation over all indices in the given interval $i$ to $k$. % and $[\cdot; \cdot]$ the concatenation of its given elements. 
	We introduce multiple layers to our approach. This was inspired by TCNs and the use of multiple layers in neural networks at large \cite{Goodfellow-et-al-2016}, we introduce multiple layers to our approach. In the case of $L\in\mathbb{N}$ layers of this mapping, we propose for a layer $j > 1$
	\begin{equation}
	\label{eq:j_s}
	{}_js_i^{(t)} = {}_{j-1}s_{i}^{(t)} \ \cdot \ {}_{j-1}s_{i-1}^{(t)} = \hat{x}_{i} \ \cdot \ \hat{x}_{i-1}.
	\end{equation}
	
\begin{figure}[htb]
		\centering
		\includegraphics[trim=0 0 0 0, clip, height=.3\textheight]{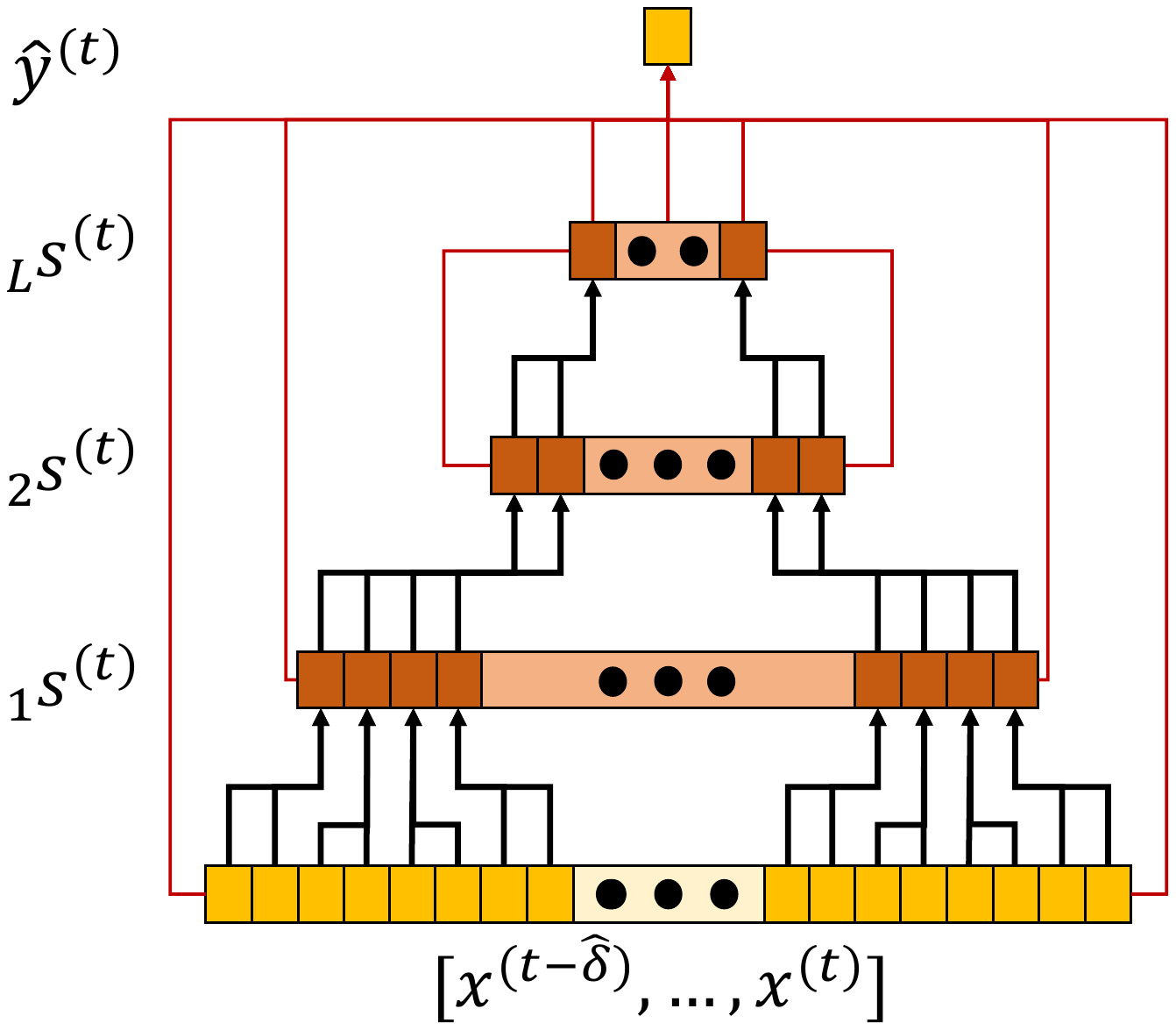}
		\caption{Proposed TCRC architecture, where black arrows ({\color{black}$\rightarrow$}) refer to the multiplied tokens and red arrows ({\color{red}$\rightarrow$}) refer to the learned mapping from the state space of each layer ${}_ls^{(t)}$ to the output $\hat{y}^{(t)}$. }%\hl{bitte pruefen, ob noch korrekt} }
		\label{fig:TC_pure}
	\end{figure}

	%Due to the iterative nature of this computation, no padding is needed as typical for classic convolutions to preserve a certain output dimensionality \cite{lecun1989generalization, pandey2022direct}.
	For mapping an accumulated state space over each layer to the output, we propose for the final state of the TCRC at the time step $t$ for ${}_js^{(t)}$ analogous to Eq.~\ref{eq:first_TC_state}: 
	\begin{equation}\label{eq:state_TC}
	s^{(t)} =f([\mdoubleplus_{i=1}^{\hat{\delta}}x^{(t-1)};\mdoubleplus_{j=1}^{L}{}_js^{(t)})],
	\end{equation}
	for $L$ layers as shown in Fig.~\ref{fig:TC_pure} and the activation function $f(\cdot)$ chosen as $\tanh(\cdot)$, with $[\cdot; \cdot]$ being the concatenation of its given elements. In conclusion, our proposed approach results in the set of hyper-parameters $\Omega' = \{\hat{\delta}, L, S_T, \beta\}$, with $|\Omega'| < |\Omega|$ and $|\cdot|$ being the cardinality of the set. 
	We allow for a zero padding, increasing the size of the state space, for potential noise as an alternative method of regularization as proposed in \cite{lukovsevivcius2009reservoir}.

	\subsection{TC-Extreme Layer (TCRC-ELM)}\label{sec:ELMlayer}
\begin{figure}[htb]
		\centering
		\includegraphics[trim=0 0 0 0,clip,height=.3\textheight]{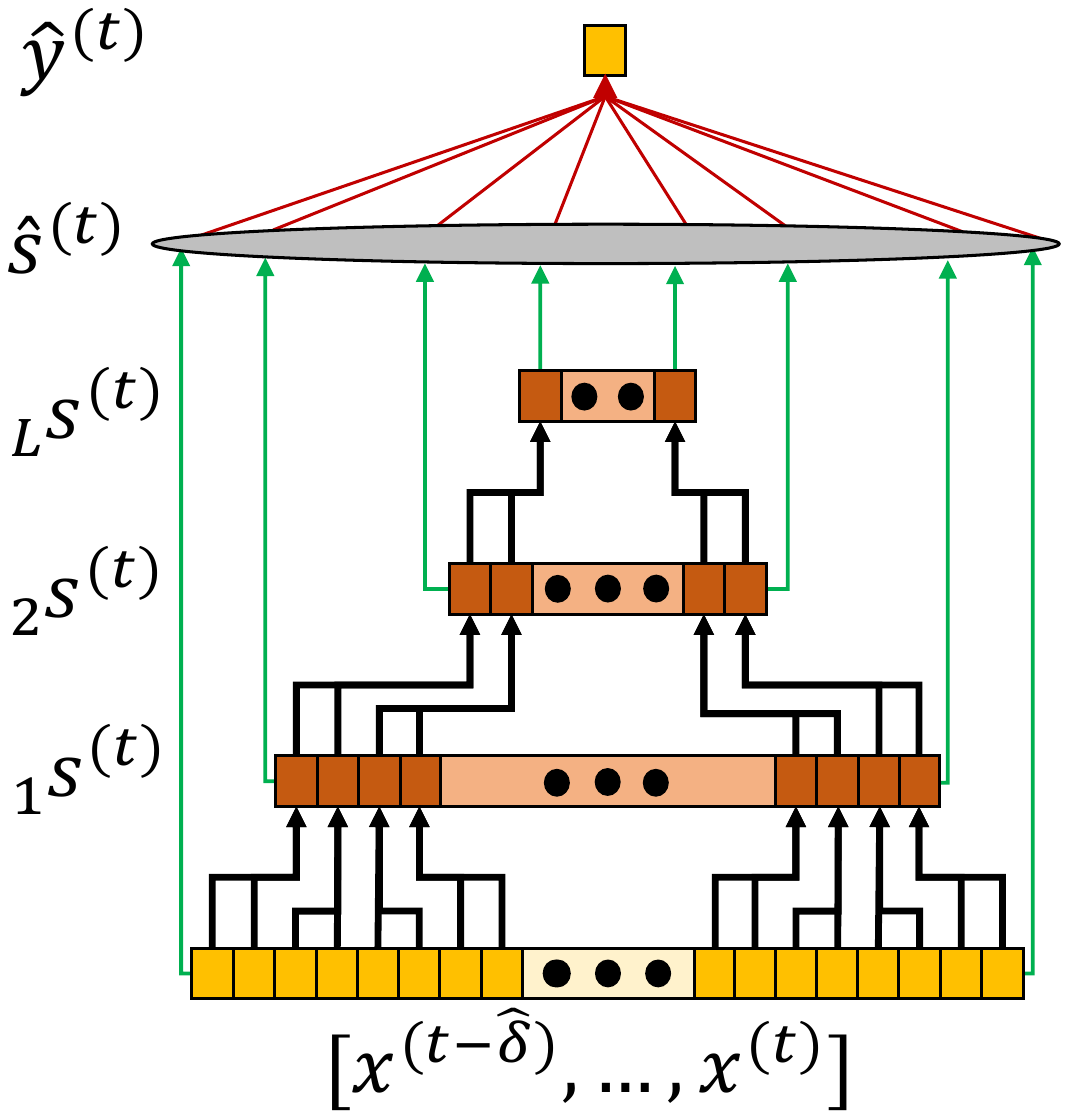}
		\caption{Proposed TCRC-ELM architecture, where black arrows ({\color{black}$\rightarrow$}) refer to the multiplied tokens, green arrows ({\color{OliveGreen}$\rightarrow$}) refer to the randomly drawn weights of $W^{\mathrm{in}}$, and red arrows ({\color{red}$\rightarrow$}) refer to the learned mapping from the state space $\hat{s}^{(t)}$ to the output $\hat{y}^{(t)}$. For the sake of simplicity we have not shown the optional learned connections from all ${}_j{s}^{(t)}$ to $\hat{y}^{(t)}$.} 
		\label{fig:TC_ELM}
	\end{figure}

	For combining the temporal convolution concept with Extreme Learning M;achines (ELM) \cite{huang2004extreme}, we propose to use a single randomly generated weight matrix and the state $s^{(t)}$ of the TCRC as input to an ELM (cp.~Fig.~\ref{fig:TC_ELM}), similar to the work of \cite{gallicchio2011architectural}. We refer to this approach as TCRC-ELM. An input layer $W^{\mathrm{in}}\in\mathbb{R}^{N^{\mathrm{r'}}\times N^{\mathrm{r}}}$ is drawn from a uniform distribution. We assume that $W^{\mathrm{in}}\sim\mathcal{U}(-\delta,\delta)$ with $\delta = \frac{1}{2}$ and, furthermore, we assume that $s^{(t)}\in\mathbb{R}^{N^\mathrm{r}}$, $\hat{s}^{(t)}\in\mathbb{R}^{N^\mathrm{r'}}$ and $N^{\mathrm{r'}}=n\cdot N^{\mathrm{r}}$ with $n, \ N^{\mathrm{r'}}\in\mathbb{N}$. To compute the additional state based on the ELM-like layer, we denote 
	\begin{equation}\label{eq:ELM_TC}
	\hat{s}^{(t)} =\hat{f}(W^{\mathrm{in}}[\mdoubleplus_{i=1}^{\hat{\delta}}x^{(t-1)};\mdoubleplus_{j=1}^{L}{}_js^{(t)}]).
	\end{equation}
	
	We utilize $\tanh(\cdot)$ as activation function $\hat{f}(\cdot)$. In conclusion, this architecture has a slightly increased number of hyper-parameters $\Omega''=\{n, \hat{\delta}, L, S_T\}$ with $|\Omega'|<|\Omega''|<|\Omega|$. 
	
	Concluding the explanation of our proposal we want to give a more implementation focused description. For this we refer to Code \ref{code:TCRC}.
	\begin{algorithm}
		\caption{Pseudocode of the TCRC-ELM method}
		\label{code:TCRC}
		\begin{algorithmic}
			\Require $\hat{\delta}\geq 1$
			\Require $t\geq \hat{\delta}$
			\Require $L\leq \hat{\delta}$
			\State $W^{\textrm{in}}\sim\mathcal{U}(-\delta,\delta)$
			\While{$t < S_T$}
			
			\State ${}_0s^{(t)} \leftarrow \mdoubleplus_{i=0}^{\hat{\delta}}s_{i}^{(t-i)}$
			\While{$l<L$}
			\State ${}_ls^{(t)} \leftarrow \mdoubleplus_{i=1}^{\hat{\delta}-l}{}_{l-1}s_{i-1}^{(t)}{}_{l-1}s_{i}^{(t)}$
			\EndWhile
			\State ${}\hat{s}^{(t)} \leftarrow W^{\textrm{in}}\mdoubleplus_{l=0}^{L}{}_{l}s_{i}^{(t)}$
			\EndWhile
			\State $W^{\textrm{out}} \leftarrow T(f([\mdoubleplus_{t=0}^{S_T}{}\hat{s}^{(t)},\mdoubleplus_{l=0}^L\mdoubleplus_{t=0}^{S_T}{}_ls^{(t)},\mdoubleplus_{t=\hat{\delta}}^{S_T}\mdoubleplus_{i=0}^{\hat{\delta}}x^{t-i}])$) \Comment{$T(\cdot)$ is the application of the}
			\Comment{Tikhonov Regularization}
			\While{$\hat{t}\leq S_P$}
			\State ${}_0s^{(S_T+\hat{t})} \leftarrow \mdoubleplus_{i=0}^{\hat{\delta}}s_{i}^{(S_T+\hat{t}-i)}$
			\While{$l<L$}
			\State ${}_ls^{(S_T+\hat{t})} \leftarrow \mdoubleplus_{i=1}^{\hat{\delta}-l}{}_{l-1}s_{i-1}^{(S_T+\hat{t})}{}_{l-1}s_{i}^{(S_T+\hat{t})}$
			\EndWhile
			\State ${}\hat{s}^{(S_T+\hat{t})} \leftarrow W^{\textrm{in}}\mdoubleplus_{l=0}^{L}{}_{l}s_{i}^{(S_T+\hat{t})}$
			
			\State $\hat{y}^{(\hat{t})} \leftarrow W^{\textrm{out}}f([\hat{s}^{(S_T+\hat{t})},\ldots])$\Comment{Compare stacking $W^{\textrm{out}}$}
			\State $x^{(S_T+\hat{t})} \leftarrow \hat{y}^{(\hat{t})}$
			\EndWhile
			%	\Ensure $y = x^n$
			%	\State $y \gets 1$
			%	\State $X \gets x$
			%	\State $N \gets n$
			%	\While{$N \neq 0$}
			%		\If{$N$ is even}
			%		\State $X \gets X \times X$
			%		\State $N \gets \frac{N}{2}$  \Comment{This is a comment}
			%		\ElsIf{$N$ is odd}
			%		\State $y \gets y \times X$
			%		\State $N \gets N - 1$
			%		\EndIf
			%		\EndWhile
		\end{algorithmic}
	\end{algorithm}

	% !TEX root = ../main.tex

	\section{Evaluation and Results}\label{sec:eval}
	In this section, we introduce the datasets used for evaluation as well as their preprocessing and the metrics used for evaluating the different approaches. Further, we evaluate the performance of state-of-the-art models, i.e. ESN, GRU and NG-RC, to establish a baseline on the selected datasets.
	Subsequently, we compare the performance of shallow TCRC and TCRC-ELM models to optimized ESN, DeepESN and GRU before also discussing results of deep TCRC architectures. We evaluate the complexity of our approach by the measured wall clock time and memory allocation at the end of the section.
	\begin{figure*}[htb]
		\centering
		\caption*{Non-chaotic variation of the Mackey-Glass time series}
		\begin{subfigure}{.32\textwidth}
			\includegraphics[trim=30 25 23 23, clip,width=\textwidth]{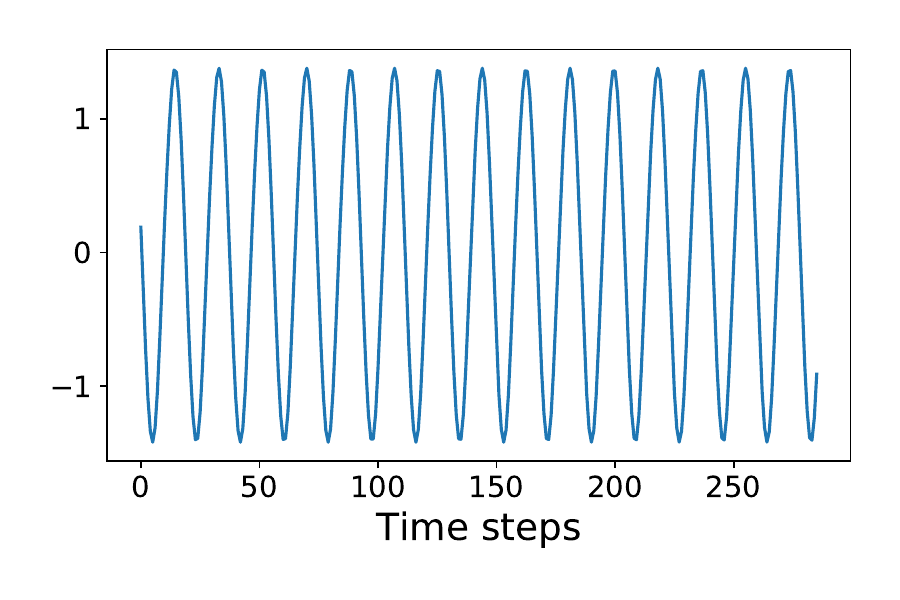}
			\caption{$\tau=5$}
		\end{subfigure}
		\begin{subfigure}{.32\textwidth}
			\includegraphics[trim=30 25 23 23, clip,width=\textwidth]{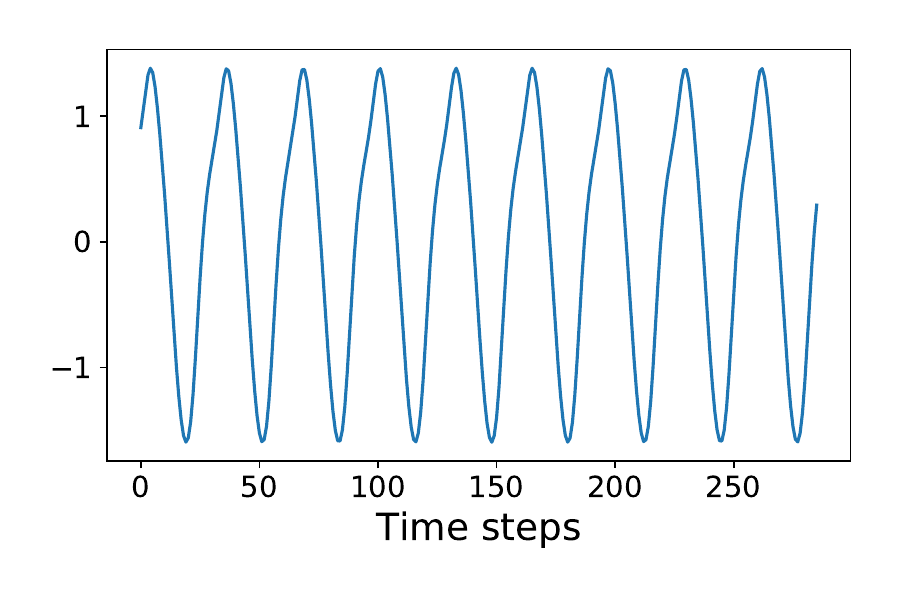}
			\caption{$\tau=10$}
		\end{subfigure}
		\begin{subfigure}{.32\textwidth}
			\includegraphics[trim=30 25 23 23, clip,width=\textwidth]{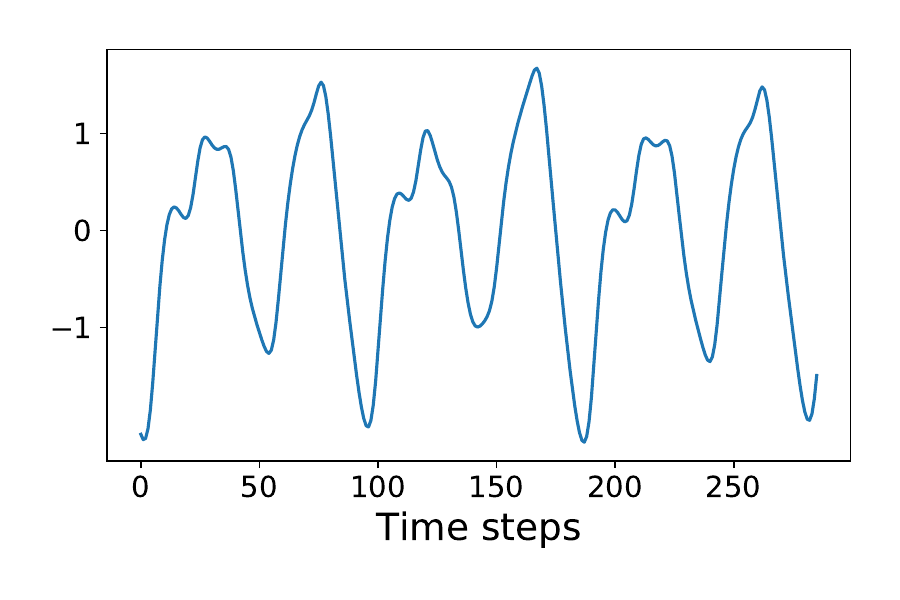}
			\caption{$\tau=15$}
		\end{subfigure}
		\caption*{Chaotic variation of the Mackey-Glass time series}
		\begin{subfigure}{.32\textwidth}
			\includegraphics[trim=30 25 23 23, clip,width=\textwidth]{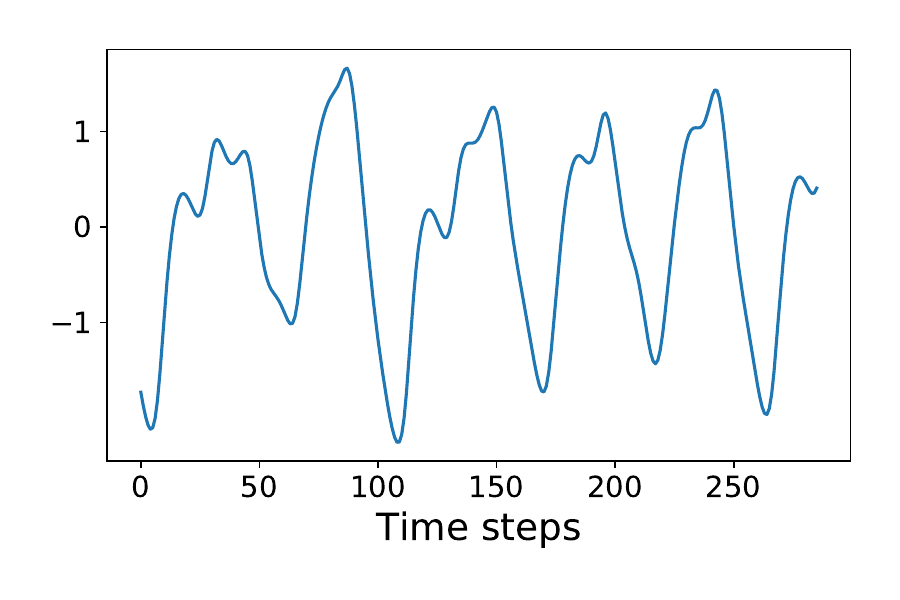}
			\caption{$\tau=17$}
		\end{subfigure}
		\begin{subfigure}{.32\textwidth}
			\includegraphics[trim=30 25 23 23, clip,width=\textwidth]{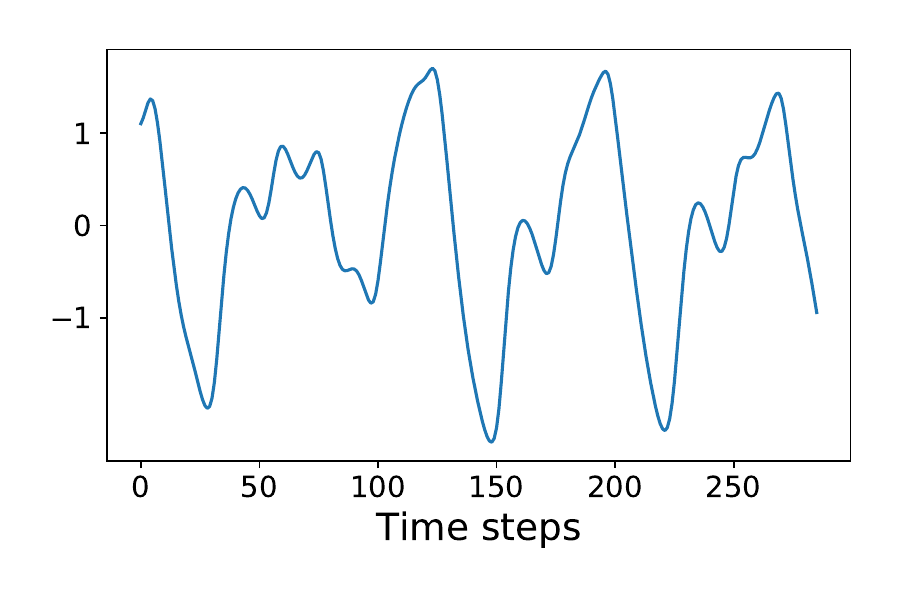}
			\caption{$\tau=20$}
		\end{subfigure}
		\begin{subfigure}{.32\textwidth}
			\includegraphics[trim=30 25 23 23, clip,width=\textwidth]{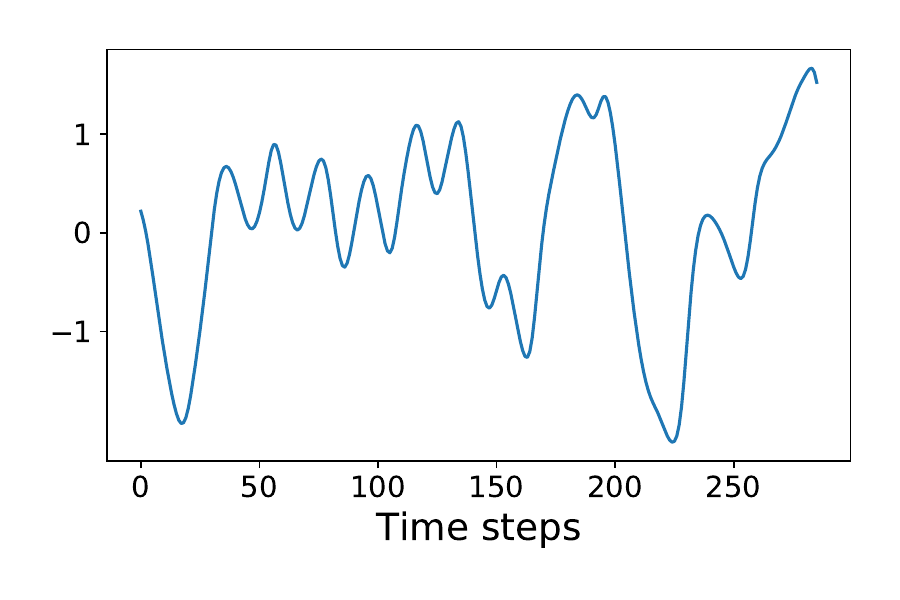}
			\caption{$\tau=25$}
		\end{subfigure}
		\caption{Visualization of the Mackey-Glass equation for $286$ time steps and different values of $\tau$, shifting from non-chaotic to chaotic with $\tau\geq17$. }% \hl{influencing the chaoticity of the time series -- korrekt???}.}
		\label{fig:mglass}
	\end{figure*}
	
	\paragraph{Dataset}
	For the evaluation of the proposed methods, we chose the Mackey-Glass equation \cite{glass2010mackey}. The selection of this dataset is motivated by its wide use in the RC literature to test ESNs, up to the proposal of the architecture in \cite{jaeger2001echo} and its in some cases chaotic nature. Albeit univariate, this makes it to an example widely comparable and simple enough to explain the relations between good performing predictors and the selected parameters yet still complex enough to be in need of a sophisticated predictive approach. We approximated the time series with the delays $\tau \in \{5,10,15,17,20,25\}$ by solving
	\begin{equation}\label{eq:MG}
	\dot{x}^{(t)} = \frac{\beta_{MG}\ x^{(t-\tau)}}{\Theta^n+x^{{(t-\tau)}^n}}-\gamma x^{(t)},
	\end{equation}
	with $[\beta_{MG}, \Theta, \gamma, n ] = [0.2, 1, 0.1, 10]$ (cp. Fig.~\ref{fig:mglass}). The parameter $\tau$ defines the number of time steps in the past, which still have to be held in memory to calculate the current time step. The figure shows how the time series become more complex with an increasing $\tau$. Thereby, $\tau = 17$ is the smallest delay resulting in chaotic behavior as explained in \cite{moon2021hierarchical}. We train with $S_T=2000$ time steps to be comparable with, e.g. \cite{VIEHWEG2022}. For the test, we have chosen $S_P=286$ time steps, representing  two Lyapunov times for the $\tau=17$ parametrization \cite{farmer1982chaotic}. 
	The approximation was done with an adaptation of the Euler method, with a step width of one. 
	As secondary dataset we utilize the SantaFe Laser time series \cite{Laser}, as well-established benchmark dataset in the field of time series forecasting. The data points were measured with the time series showing chaotic behaviour, for explanation and further discussion we refer to the work of Weiss et. al \cite{weiss1995lorenz}. We have limited the prediction of this dataset to $100$ times steps according to \cite{weigend1993results}.
	We normalize all datasets to a zero mean value $\mu(\cdot)$ and a standard deviation $\hat{\sigma}(\cdot)$ of one. This is motivated by the utilized $\tanh(\cdot)$ non-linearity which saturates at values beyond the $[-1,1]$ range. With the input being given to the output layer with a residual connection this limits the variance in the states, reducing the expressiveness of the $s^{(\cdot)}$ and $\hat{s}^{(\cdot)}$ respectively for the learning task. For the benefits of normalization, we also reference to \cite{Goodfellow-et-al-2016} and to empirical observations such as those of \cite{teutsch2023data}. We normalized the input as well as the targets by use of the z-score \cite{carey2010t}:
	\begin{equation}\label{eq:zscore}
	X = \frac{X_0-\mu(X_0)}{\hat{\sigma}(X_0)}.
	\end{equation}
	Here we use $X_0$ as the original input (cp. Eq.~\ref{eq:MG}), i.e. because of the characteristic of all time series to have values in a fixed range, all time steps computed. $X\in\mathbb{R}^{N^{\mathrm{in}}\times ({S_I+S_T})}$ is the matrix from which the inputs are drawn. The target outputs are drawn from $Y,Y_0\in\mathbb{R}^{N^{\mathrm{out}}\times {(S_I+S_T+S_P)}}$ analogous to the inputs (cp. Eq.~\ref{eq:zscore}).

	\paragraph{Training and Evaluation}
	We limit the shown results to the tested set of hyper-parameters, i.e. the best performing set for our approaches. For the ESN and TCRC-ELM, which use randomness in their initialization, we report the mean error over $15$ runs. In regard to our own approach we tested multiple delays $\hat{\delta}$ of each tested $\tau$ in the range $\hat{\delta}\in[2,1900]$, substantially larger than any potential $\tau$ used for the Mackey-Glass equations. As training objective and for comparison of trained models, we use the mean squared error $\mathrm{MSE}$
	\begin{equation}\label{eq:MSE}
	\mathcal{L}=\frac{\sum_{t'=0}^{S_P}(y^{(t')}-\hat{y}^{(t')})^2}{S_P}. 
	\end{equation}
	
	For the evaluation itself, we test the TCRC and TCRC-ELM with only the regularization parameter $\beta$ and the delay $\hat{\delta}$ varied at first. In a second step we show the influence of allowing for a deep TCRC, using multiple layers as shown in Figure~\ref{fig:TC_pure} both for the TCRC itself as well as the TCRC-ELM. In both evaluation steps, we compare them to an ESN with optimized hyper-parameters as well as an GRU with optimized input size. We additionally report the number of valid time steps, whereby we have chosen a time steps as valid, for as long as it stays in an absolute difference of below one standard deviation of the data around the ground truth, reasoned by it staying inside the orbit of the attractor. 
	\paragraph{Parametrization}
	The baseline for the ESN was the default setting proposed by \cite{VIEHWEG2022}, from which we used a grid search approach with $N^{\textrm{res}}\in[1000, 3000]$, $\rho\in[0.5,1.5]$, density $\sigma(W^{\mathrm{r}})\in [0.1,1]$as well as the regularization coefficient $\beta\in[10^{-10},10^{-3}]$. Other approaches to optimize the parameter selection are possible as shown, e.g. by \cite{soltani2023echo, zhai2023model} with a comparison of optimization methods used in literature. 
	In case of the AEESN we started from the parametrization utilized in the proposing literature of \cite{ma2020deepr} and a grid search. For our own proposal we used a grid search approach with a delay $\hat{\delta}\in[1,1900]$ time steps, the number of layers $l\in[1,6]$ and a regularization coefficient $\beta\in[10^{-6},1]$.
	For the GRU we utilized a grid search with a state space size from four and so smaller than our smallest delay up to $2048$, a history and input of up to $150$ time steps and allowing for a number of layers $l\in[1,5]$, while training for a number of epochs $e\in[100,2000]$ with the Adam optimizer \cite{kingma2014adam, paszke2017automatic}.

	%\paragraph{Baselines}
	\subsection{Baselines}
	In this section we discuss the established models for time series prediction used in this work. They are used here as baseline to measure the results of our approaches TCRC and TCRC-ELM against.
		\begin{table*}
		\centering
		\caption{MSE of best performing Basic ESN, AEESN, GRU and TCRC for prediction of $286$ time steps for different $\tau$ of the Mackey-Glass equation and $100$ time steps for the Santa Fe Laser dataset. We present the best tested $\rho\in[0,1.5]$ and $N^{\mathrm{r}}\in\{1000,2000,3000\}$ as well as best tested input lengths $\hat{\delta}$ for the GRU and TCRC. The green values show a decrease while an increase is marked by red values which relative to the ESN for each dataset. The best performing model is marked in bold. Additional regularization is not used.}
		\begin{adjustbox}{width=\textwidth}
			\begin{tabular} {c|c|c|c|lr|lr}
				\toprule
				\multicolumn{1}{>{\centering\arraybackslash}m{.192\textwidth}|}{\textbf{Dataset} } &\multicolumn{1}{>{\centering\arraybackslash}m{.192\textwidth}|}{\textbf{ESN} } &\multicolumn{1}{>{\centering\arraybackslash}m{.192\textwidth}|}{\textbf{AEESN} }  &\multicolumn{1}{>{\centering\arraybackslash}m{.192\textwidth}|}{\textbf{GRU} } &\multicolumn{2}{>{\centering\arraybackslash}m{.192\textwidth}|}{\textbf{TCRC} } &\multicolumn{2}{>{\centering\arraybackslash}m{.192\textwidth}}{\textbf{TCRC-ELM} }
				\\ 
				\midrule
				\cellcolor[gray]{0.8}{\textbf{Mackey-Glass, $\mathbf{\tau=}$}}
				& \cellcolor[gray]{0.8}& \cellcolor[gray]{0.8}& \cellcolor[gray]{0.8}& \cellcolor[gray]{0.8}MSE& \cellcolor[gray]{0.8}rel. impr.& \cellcolor[gray]{0.8}MSE & \cellcolor[gray]{0.8}rel. impr.\\
				\midrule
				{$5$}& {$5.72\cdot10^{-3}$}&$2.37\cdot10^{1}$&$1.90\cdot10^{-4}$&$\boldsymbol{1.25\cdot10^{-6}}$ &$\boldsymbol{{\color{OliveGreen}{-99.98\%}}}$&$4.50\cdot10^{-6}$&$\color{OliveGreen}{-99.92\%}$\\
				\midrule
				{$10$} & {$3.59\cdot10^{-2}$} &$1.26\cdot10^{-1}$&$7.61\cdot10^{-3}$& $\boldsymbol{{2.31\cdot10^{-9}}}$ &$\boldsymbol{{{\color{OliveGreen}{-99.99\%}}}}$&$2.25\cdot10^{-7}$&$\color{OliveGreen}{-99.99\%}$\\
				\midrule
				{$15$} & {$5.41\cdot10^{-2}$} &$2.45\cdot10^{-2}$&$2.50\cdot10^{-1}$&$\boldsymbol{4.79\cdot10^{-4}}$ &$\boldsymbol{{\color{OliveGreen}{-99.11\%}}}$ &$7.36\cdot10^{-3}$&$\color{OliveGreen}{-86.40\%}$\\
				\midrule
				{$17$} & {$2.99\cdot10^{-1}$} &$7.95\cdot10^{-2}$&${4.69\cdot10^{-1}}$&$9.33\cdot10^{-1}$ &${\color{red}{+212.04\%}}$ &$\boldsymbol{4.68\cdot10^{-2}}$&$\boldsymbol{{\color{OliveGreen}{-84.35\%}}}$\\
				\midrule
				{$20$} & $\boldsymbol{3.20\cdot10^{-1}}$ &$3.88\cdot10^{-1}$&$6.10\cdot10^{-1}$& ${5.58\cdot10^{-1}}$ &${{\color{red}{+74.38\%}}}$ &$3.24\cdot10^{-1}$&${{\color{red}{+1.25\%}}}$\\
				\midrule
				{$25$} & {$1.06$}  &$\boldsymbol{1.70\cdot10^{-1}}$&${8.15\cdot10^{-1}}$&$1.25$ &${\color{red}{+17.92\%}}$&$5.33\cdot10^{-1}$&$\color{OliveGreen}{-49.72\%}$\\
				\midrule
				\cellcolor[gray]{0.8}{\textbf{Laser} }& \cellcolor[gray]{0.8}& \cellcolor[gray]{0.8}& \cellcolor[gray]{0.8}& \cellcolor[gray]{0.8}& \cellcolor[gray]{0.8}& \cellcolor[gray]{0.8}& \cellcolor[gray]{0.8}\\
				\midrule
				&\boldsymbol{{$6.31\cdot10^{-2}$}}&$9.13\cdot10^{0}$&$5.92\cdot10^{-1}$&${4.33\cdot10^{-1}}$ &${\color{red}{+581.21\%}}$&$9.48\cdot10^{-1}$&$\color{red}{+1402.38\%}$\\
				\bottomrule
			\end{tabular}
		\end{adjustbox}
		\label{tab:esn_GRU}
	\end{table*} 
	\paragraph{Echo State Network (ESN)}
	We use a standard Echo State Network (ESN) to establish a baseline for our evaluation on the generated datasets (cp.~Tab.~\ref{tab:esn_GRU}). For the ESN, we have tested spectral radii from $\rho = 0$, basically resulting in an ELM since the ESN's memory has no effect for that case, up to $\rho=1.5$ and state space sizes $N^{\mathrm{r}}\in\{1000; \ 2000; \ 3000\}$ with a connectivity of $\sigma(W^{\mathrm{r}})$ between $0.1$ and $1.0$ and no leakage rate applied. 
	We observe the error to increase together with $\tau$ as to be expected, from $5.72\cdot10^{-3}$ to $1.06$. The ESN is the only predictor with a substantial increase between the two lowest values of $\tau$ while all other predictors are comparable for this values (GRU) or even decrease from $\tau=5$ to $\tau=10$. A substantial increase we observe between $\tau=15$ and $\tau=17$. We argue that this is expected due to the Mackey-Glass equation switching between those values from non-chaotic to chaotic behavior. The predictive capabilities of the ESN are insufficient in the bounds of hyper-parameters, chosen to be comparable with our proposed approach for the most chaotic dataset with an error above one for values normalized to a standard deviation of one. The deviation across multiple random initialization led to the observation of the ESNs prediction varying by an order of magnitude between the runs, albeit only for the chaotic time series. The ESN achieved the best prediction for the SantaFe Laser dataset, one order of magnitude better compared to the proposed approaches and other baselines.

	\paragraph{Gated Recurrent Unit (GRU)}
	We tested a Gated Recurrent Unit (GRU) with a single up to five layers to be comparable to the ESN, while still allowing for the advantages of a deep neural network, with an input length $\hat{\delta}\in[1,150]$ and a size of the GRU cell of up to $2048$. We observe the prediction of a fixed value with a single time step instead of showing dynamic behavior. This seems surprising but is comparable to results in the literature \cite{shahi2022prediction}. We observe a benefit of increasing the input by use of delays across multiple time steps as shown in Tab.~\ref{tab:esn_GRU}. Hereby we tested delays up to $\hat{\delta}=150$, larger than in some literature, e.g. \cite{teutsch2022flipped}but still observed the best performing, and here reported GRU to utilize a single layer.
	The error increases with an increase of $\tau$. % with one exception of $\tau=25$ but does not change as substantially as the ESN by switching from non-chaotic to chaotic datasets. 
	In comparison to the ESN the single layered GRU is better by up to $78.80\%$ for the non-chaotic time series of $\tau=10$. % and $78.80\%$ for the chaotic variation with $\tau=25$.
	Using the GRU results in a worse prediction for $\tau=15$ to $\tau=20$, while in a better one for $\tau=25$ and also the hyperchaotic parametrization of $\tau=30$, shown in Tab. \ref{tab:app:AEESN}. % by $4.58$ and $3.00$ times. For $\tau=25$ the GRU yields the overall best predictive result with $1.23\cdot10^{-1}$. 
	As shown in Tab. \ref{tab:app:valsteps} the GRU can achieve a valid prediction of up to the full $286$ time steps with missing the prediction for $\tau=15$. We argue for this prediction to be an outlier, our approach to the approximation of the dataset being to inaccurate, leading to a behaviour, more shifted towards chaotic than represented in the training data.
	For the SantaFe Laser time series we observe the GRU to be a better predictor than the AEESN and TCRC-ELM, but worse than TCRC and ESN.
	\paragraph{Next Generation Reservoir Computing (NG-RC)}
	Next Generation Reservoir Computing (NG-RC) is described by its authors as comparable to ESNs with hundreds to thousands of nodes \cite{gauthier2021next}. Therefore, we consider the evaluated reservoir sizes for ESN in our study to be a valid comparison. For the lowest tested value of $\tau=5$ we observe the best prediction with a delay of $\hat{\delta} = 10$,  the error decreases by above $99.99\%$. In case of $\tau = 10$ the selection of $\hat{\delta}=10$ compared to the ESN leads to a decrease of $64.90\%$. The best MSE for $\tau\geq15$ is observed for a very low delay $\hat{\delta}=2$ which leads to a constant prediction. An increase of the $\hat{\delta}$ leads to the network seemingly being unable to learn. Due to this, we do not further use the NG-RC for comparisons and do not discuss the prediction of the chaotic time series. We observe a trend to a large $N^{\mathrm{in}}$ with $\tau<<N^{\mathrm{in}}$. Also we observe, compared to the ESN, a more drastic increase of the MSE from non-chaotic to chaotic variants of the system.
	We observe a similar result of not being close to the targeted trajectory for the SantaFe laser dataset, showing the need of an alternate non-linear function for this time series, than e.g. the Lorenz'63 attractor of \cite{lorenz1963deterministic} utilized in \cite{gauthier2021next}.
	\paragraph{DeepR-ESN (AEESN)}
	As further baselines we used the Deep ESN with Autoencoders between the reservoirs (AEESN) proposed by \cite{ma2020deepr}. We found the AEESN for our preprocessing of our datasets and chosen time series to vary between results substantially with in some cases, especially for $\tau=5$ with an error of up to more than $225$, which results in such an incomparable large mean reported in Tab. \ref{tab:TCRCs}. The error for $\tau=30$ is reported in the appendix (cp. Tab.~\ref{tab:app:AEESN}). Reporting only the best initializations, e.g. in Tab. \ref{tab:TCRCs} and Tab. \ref{tab:app:valsteps} shows the AEESN as beneficial compared to the ESN in some cases, even when measured in the number of valid time steps of the prediction. We observe it to be superior to both of our proposed approaches for $\tau=25$, outperforming the TCRC for all chaotic time variations of Mackey-Glass.
	
	In case of the Santa Fe Laser time series the AEESN is the worst performing baseline, as shown in Tab. \ref{tab:TCRCs} and \ref{tab:esn_GRU} because of the utilization of the mean across $15$ initializations.
	\begin{table*}[htb]
		\centering
		\caption{MSE of best performing configurations of TCRC and TCRC-ELM as well as best baseline for prediction of $286$ time steps for different $\tau$ of the Mackey-Glass equation as well as $100$ time steps for the SantaFe Laser dataset. We present the best tested input lengths $\hat{\delta}$ for our TCRC and TCRC-ELM with the best tested number of layers $L$. The green values show a decrease while an increase is marked by red values relative to the best baseline for each dataset. The best performing model is marked in bold.
		} 
		%\begin{adjustbox}{width=\textwidth}
		
		\begin{tabularx}{.95\textwidth}{c|c | c | clr | clr }
			\toprule
			\multicolumn{2}{c|}{\textbf{Dataset}}& \multicolumn{1}{c|}{\textbf{Baseline}}  & \multicolumn{3}{c|}{\textbf{TCRC} } &\multicolumn{3}{c}{\textbf{TCRC-ELM} } \\ 
			\midrule
			\cellcolor[gray]{0.8}{\textbf{Mackey-Glass}}&\cellcolor[gray]{0.8}$\boldsymbol{\tau}$&\cellcolor[gray]{0.8} &\cellcolor[gray]{0.8}$L$&\cellcolor[gray]{0.8}MSE&\cellcolor[gray]{0.8}rel. impr.&\cellcolor[gray]{0.8}$L$&\cellcolor[gray]{0.8}MSE &\cellcolor[gray]{0.8}rel. impr.\\
			\midrule
			\multirow{4}{*}{non-chaotic}&{$5$}& {$1.90\cdot10^{-4}$}&4&$\boldsymbol{1.15\cdot10^{-6}}$&$\boldsymbol{{\color{OliveGreen}{-99.39\%}}}$&2&$2.81\cdot10^{-6}$&${{\color{OliveGreen}{-98.52\%}}}$\\ 
			\cmidrule{2-9}
			&{$10$} & {$7.61\cdot10^{-3}$} &1&$\boldsymbol{{2.31\cdot10^{-9}}}$ &$\boldsymbol{{\color{OliveGreen}{-99.99\%}}}$&1&$1.37\cdot10^{-7}$&$\color{OliveGreen}{-99.99\%}$\\ 
			\cmidrule{2-9}
			&{$15$} 
			& {$2.45\cdot10^{-3}$} &3&$\boldsymbol{4.07\cdot10^{-4}}$ &$\boldsymbol{{\color{OliveGreen}{-98.38\%}}}$&1&${2.22\cdot10^{-3}}$ &${{\color{OliveGreen}{-90.94\%}}}$\\ 
			\midrule
			\multirow{4}{*}{chaotic}&{$17$} 
			& {$7.95\cdot10^{-2}$} &3&$2.22\cdot10^{-1}$ &${\color{red}{+79.25\%}}$&1&$\boldsymbol{4.35\cdot10^{-2}}$ &$\boldsymbol{{\color{OliveGreen}{-45.28\%}}}$\\
			\cmidrule{2-9}
			&{$20$} & $3.20\cdot10^{-1}$ &4&${1.70\cdot10^{-1}}$ &${{\color{OliveGreen}{-46.88\%}}}$&1&$\boldsymbol{{1.16\cdot10^{-1}}}$ &$\boldsymbol{{{\color{OliveGreen}{-63.75\%}}}}$\\ 
			\cmidrule{2-9}
			&{$25$} 
			& \boldsymbol{$1.70\cdot10^{-1}$} &1& $5.67\cdot10^{-1}$ &${\color{red}{+233.53\%}}$&1&${3.98\cdot10^{-1}}$ &${{\color{red}{+134.12\%}}}$\\ 
			\midrule
			\cellcolor[gray]{0.8}{\textbf{Laser}}&\cellcolor[gray]{0.8}&\cellcolor[gray]{0.8}&\cellcolor[gray]{0.8}&\cellcolor[gray]{0.8}&\cellcolor[gray]{0.8}&\cellcolor[gray]{0.8}&\cellcolor[gray]{0.8}&\cellcolor[gray]{0.8}\\
			\midrule
			&&\boldsymbol{{$6.31\cdot10^{-2}$}}&$1$&${4.33\cdot10^{-1}}$ &${\color{red}{+581.46\%}}$&$1$&$9.48\cdot10^{-1}$&$\color{red}{+1402.38\%}$\\
			\bottomrule
		\end{tabularx}
		%\end{adjustbox}
		
		\label{tab:TCRCs}
	\end{table*} 
	%\begin{table*}
	%	\centering
	%	\caption{MSE of best performing Basic ESN, AEESN, GRU and TCRC for prediction of $100$ time steps for the SantaFe Laser. We present the best tested $\rho\in[0,1.5]$ and $N^{\mathrm{r}}\in\{1000,2000,3000\}$ as well as best tested input lengths $\hat{\delta}$ for the GRU and TCRC. The green values show a decrease while an increase is marked by red values which relative to the ESN for each given $\tau$. The best performing model is marked in bold}
	%	\begin{adjustbox}{width=\textwidth}
	%		\begin{tabular} {c|c|c|lr|lr}
	%			\toprule
	%			\multicolumn{1}{>{\centering\arraybackslash}m{.192\textwidth}|}{\textbf{ESN} } &\multicolumn{1}{>{\centering\arraybackslash}m{.192\textwidth}|}{\textbf{AEESN} }  &\multicolumn{1}{>{\centering\arraybackslash}m{.192\textwidth}|}{\textbf{GRU} } &\multicolumn{2}{>{\centering\arraybackslash}m{.192\textwidth}|}{\textbf{TCRC} }  		&\multicolumn{2}{>{\centering\arraybackslash}m{.192\textwidth}}{\textbf{TCRC-ELM} }  	
	%			\\ 
	%			&&&MSE&rel. impr.&MSE &rel. impr.\\
	%			\midrule
	%			\boldsymbol{{$6.31\cdot10^{-2}$}}&$9.13\cdot10^{0}$&$5.92\cdot10^{-1}$&${4.33\cdot10^{-1}}$ &${\color{red}{+681.46\%}}$&$9.48\cdot10^{-1}$&$\color{red}{+1402.38\%}$\\
	
	%			\bottomrule
	%		\end{tabular}
	%	\end{adjustbox}
	%	\label{tab:laser}
	%\end{table*} 
	\subsection{Proposed approaches}
	Here we want to discuss both of our approaches. We start by discussion of the the shallow TCRC and TCRC-ELM. After this the influence of multiple layers in the TCRC and subsequently the TCRC-ELM will be evaluated.
	
	\paragraph{TCRC}
	
	For the TCRC the predictions show a trend to an increased error from $\tau = 5$ to $\tau = 25$ as shown in \ref{tab:esn_GRU} for the default set of hyper-parameters, using a single layer. Unintuitive we argue seems the decrease of the error from $\tau = 5$ to $\tau=10$. We argue this to be due to the periodicity of both time series with $\tau=10$ evolving slower and showing less dynamics in $286$ time steps than $\tau=5$. Overall we observe the TCRC to result in the best predictions for the non-chaotic variations of the Mackey-Glass equation up to $\tau\leq15$. Thereby, the error compared to the ESN is reduced by up to $99.99\%$ for $\tau=10$.  For all non-chaotic time series we observe a reduction of the error of over $99.10\%$. For the chaotic variations we observe the TCRC with a default set of hyper-parameters to be an inferior predictor compared to the ESN and AEESN. Especially for $\tau=17$ the error is increased in comparison to the ESN by $212.04\%$. With a further increase in $\tau$ the relative decrease in performance declines to $74.38\%$ and $17.92\%$ respectively, but we argue this to be because of the worsening performance of the ESN. 
	In regard to the number of valid time steps we have shown in Tab. \ref{tab:app:valsteps} that our proposed approach is able to achieve a valid prediction for the complete interval of our test data, staying in the attractor, for the non-chaotic time series. The outlier of $\tau=20$ is repeated not only in regard to the MSE but also the number of valid time steps, showing a better prediction than for $\tau=17$. For the most chaotic tested time series, outside hyperchaos, the TCRC still achieves $100$ time steps, before the shift in the timing is too large, albeit still being inside the value space of the attractor, as shown in Fig. \ref{fig:app:pred_plots_TCRC}.
	
	We observe the TCRC to be the second best predictor in case of the Santa Fe Laser dataset, being $26.86\%$ better than the third best performing baseline of GRU, as shown in Tab. \ref{tab:TCRCs} and Tab. \ref{tab:esn_GRU}, but worse than the ESN.
	\paragraph{TCRC-ELM}
	For a single layer, the TCRC-ELM yields superior results, similar to the TCRC, in comparison to the ESN for all datasets with the exception of $\tau = 20$ (cp. Tab.~\ref{tab:esn_GRU}). 
	Like for the TCRC, the error decreases from $\tau=5$ to $\tau=10$ by $95.00\%$. For the %\hl{still -- warum noch???}
	non-chaotic variations, up to including %\hl{with ODER including??}
	$\tau=15$, we observe a substantial increase in the error compared to the TCRC, albeit still being better than the ESN, GRU and AEESN. For the least chaotic time series, we observe the TCRC-ELM to be the overall best tested predictor. The variation of $\tau=20$ is the only case in which the shallow TCRC-ELM yields $1.25\%$ worse results than the ESN. For the most chaotic time series with $\tau=25$, the TCRC-ELM is the second best performing predictor, after the AEESN. 
	We see, that the TCRC-ELM achieves the same number of valid time steps as the TCRC for the non-chaotic time series, while outperforming it for the chaotic ones, as shown in Tab. \ref{tab:app:valsteps} and exemplary visualized in Fig. \ref{fig:app:pred_plots_TCRCELM}.
	The TCRC-ELM is the second worst performing predictor for the Santa Fe Laser dataset in our tests, drifting away from the dynamics of the trajectory, increasing the error by still seemingly correct behavior.
	
	In total we argue to have managed to mitigate the negative influence of randomness, while preserving the ability to handle memory-dependent time series, as we set as a problem in Sec. \ref{sec:intro}, more specifically as our first research question. Furthermore, we have shown the ability of our proposed approach to handle chaotic time series, which we have set as our second research question.
	\begin{figure*}[htb]
		\centering
		\begin{subfigure}{.4\textwidth}
			\hspace{-.1cm}\includegraphics[width=\textwidth]{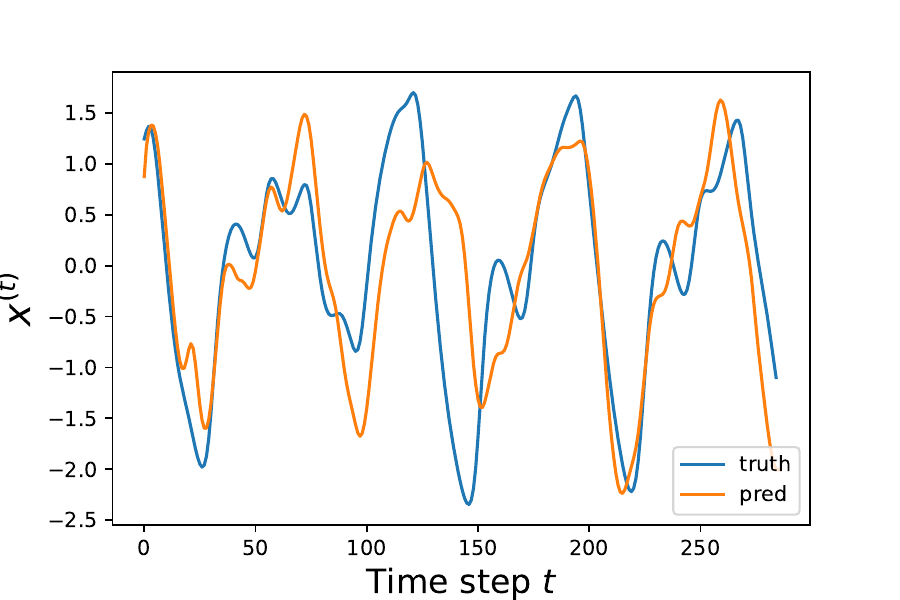}
			\caption{TCRC, $1$ Layer}
		\end{subfigure}
		\begin{subfigure}{.4\textwidth}
			\hspace{-.1cm}\includegraphics[width=\textwidth]{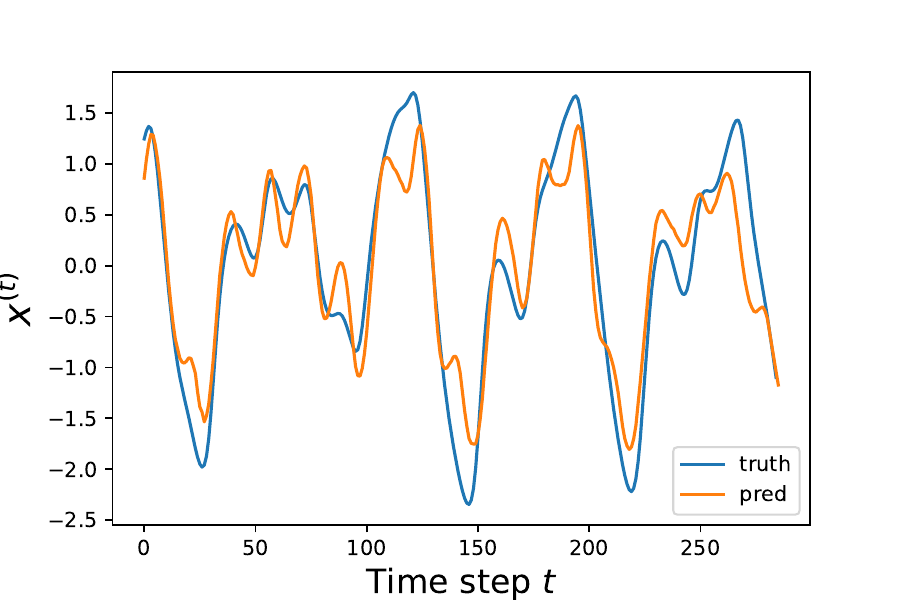}
			\caption{TCRC, $4$ Layer}
		\end{subfigure}
		\caption{Exemplary visualization of the prediction and ground truth for $\tau=20$ for the TCRC}
		\label{fig:plotted}
	\end{figure*}
	\paragraph{Multi-Layered TCRC}
	With the introduction of multiple layers and so a deep TCRC we observe a further reduction of the error (cp. Tab.~\ref{tab:TCRCs}). Even for the single layered case, we now allow for noise as additional regularization mechanism (cp. Sec.~\ref{sec:TCRC}). For $\tau=5$, we observe the lowest error for the TCRC, in this case with four layers and a reduction of $8.00\%$, compared to the shallow TCRC. The TCRC-ELM benefits from a second layer with a decrease of $37.56\%$ compared to its single layered variation. In difference to this, the time series based on $\tau=10$ does not benefit from a deep TCRC, and neither does the TCRC-ELM, which solely decreases in error due to regularization. For the most dynamic but non-chaotic time series we observe the TCRC as best performing predictor benefitting from three layers, while the TCRC-ELM benefits from the allowed noise and larger state space. For the chaotic time series, the TCRC shows clearly decreased error with multiple layers and noise outperforming the ESN for all $\tau\geq17$. In the least chaotic case the TCRC uses three layers as for $\tau=15$. For $\tau=17$ the error decreases compared to the ESN by $25.75\%$. The TCRC-ELM benefits from noise for $\tau=17$, resulting in a decrease compared to the shallow network of $7.05\%$. The time series approximated with $\tau=20$ results in a four layered TCRC and a noisy TCRC-ELM being able to outperform the ESN by $46.88\%$ and $63.75\%$ respectively. Also for $\tau=25$, our proposed approaches outperform the ESN, while being worse predictors than the AEESN, both using noise and a single layer. 
	As a difference to the TCRC, we observe no substantial benefit of multiple layers for the TCRC-ELM in this study. The best performing TCRC-ELMs uses a single layer for $\tau\geq10$ with only $\tau=5$ benefiting from two layers. In case of the hyperchaotic Mackey-Glass variation with $\tau=30$ reported in Tab. \ref{tab:app:AEESN} we observe the TCRC-ELM to be slighty better than the TCRC, similar to the behavior on chaotic datasets of Mackey-Glass with an decrease in error of $5.61\%$.
	With the introduction of the TCRC-ELM we observe a benefit of the reintroduction of randomness, allowing for better resulting the case of chaotic time series across all approximated time series, with a disadvantage in case of the Santa Fe Laser dataset.
	\subsection{Complexity}
	\begin{table*}[htb]
		\centering
		\caption{
			Runtime in seconds and maximum memory allocation in MBit of each tested model
		}
		\label{tab:runtime}
		\begin{tabular}{c|c | c|c|r | r | r  }
			\hline 
			Metric&$\simeq\boldsymbol{N^{\mathrm{r}}}$& \textbf{ESN} & \textbf{AEESN} & \textbf{GRU} &\textbf{TCRC}   &\textbf{TCRC-ELM}   \\ 
			\hline
			\hline
			\cellcolor[gray]{0.8}Runtime&\cellcolor[gray]{0.8}&\cellcolor[gray]{0.8}&\cellcolor[gray]{0.8}&\cellcolor[gray]{0.8}&\cellcolor[gray]{0.8}&\cellcolor[gray]{0.8}\\
			
			\hline												
			&{$300$}&\boldsymbol{$0.97$}&$97.29$&$10.18$& {{$6.88$}}&$5.18$\\ 
			\hline
			%{$1000$} &$--$& {$--$} &$--$&$--$&$--$\\
			%\hline 
			&{$2000$} 
			&\boldsymbol{$22.22$}&$305.48$&$59.17$&{$693.02$} &$176.92$\\% &$2.92\cdot10^{-2}$\\  
			\hline \hline
			\cellcolor[gray]{0.8}Memory Alloc.&\cellcolor[gray]{0.8}&\cellcolor[gray]{0.8}&\cellcolor[gray]{0.8}&\cellcolor[gray]{0.8}&\cellcolor[gray]{0.8}&\cellcolor[gray]{0.8}\\
			\hline
			&{$2000$}  &{$2447$}&\boldsymbol {$1424$}&$3108$&$2864$ &$2834$\\% &$2.92\cdot10^{-2}$\\  
			\hline 
			\hline
			
		\end{tabular}
	\end{table*} 
	In terms of runtime we observe an increase for a single run with a large $N^{\mathrm{r}}$ for the TCRC and TCRC-ELM as shown in Tab. \ref{tab:runtime} compared to the ESN. The smaller delay for the TCRC-ELM of the same state space size as the TCRC results in it being faster than the TCRC. The ESN is per run substantially faster, in need of only $18.73\%$ of the runtime of the TCRC-ELM in case of the smaller compared state space. In case of the larger state space the ESNs runtime increases by factor $22.91$. In case of TCRC and TCRC-ELM it increases by $100.73$ and $34.15$, respectively. We observe our proposed approaches to be faster than AEESN and GRU for the smaller state space, but scaling worse with an increase in state space size.

	In case of allocated memory the ESN is the second smallest model, saving $417$ MBit, $17.04\%$, respectively compared to the TCRC as the largest model (cp. Tab.~\ref{tab:runtime}). In comparison to the TCRC-ELM the ESN is $387$ MBit, and and such $15.82\%$ smaller. The AEESN is the overall smallest model, needing around half the memory compared to TCRC and TCRC-ELM, a reduction of $50.28\%$ and $49.75\%$ compared to TCRC and TCRC-ELM, respectively. The GRU is the most memory intensive predictor, increasing the needed memory compared to TCRC and TCRC-ELM by $8.52\%$ and $9.67\%$, respectively.

	\section{Discussion}\label{sec:discussion}
	In this work, we propose a new mapping and presented its use with multiple layers as well as delays. We observe our approach in case of optimized hyper-parameters to result in an up to $99.99\%$ and $85.45\%$ lower error than the also optimized ESN for non-chaotic and chaotic time series respectively. The introduction of the random mapping to the TCRC, resulting in the TCRC-ELM improves the predictive capabilities of the TCRC for chaotic time series but decreases it for non-chaotic time series. Still, both of our approaches decrease the error for the time series with $\tau\leq15$ by over $95.00\%$ in comparison to the ESN and by over $90.00\%$ compared to the best baseline of ESN, AEESN and GRU. For the chaotic variations, the TCRC-ELM still achieves a reduction by $85.45\%$, the TCRC up to $46.88\%$ compared to the ESN. In conclusion, our approaches outperform all baselines except AEESN for the most chaotic approximated time series $\tau=25$. We argue that this result shows that the predictive capabilities of reservoir computing can be achieved without the drawback of randomness in the weights but also shows still existing limits compared to the more learned GRU and more complex AEESN, the later of which outperforms our approaches for the hyperchaotic $\tau=30$. In case of the measured SantaFe Laser dataset the ESN outperforms both of our approaches with the TCRC being the second best predictor.
	In regard to our research questions, we have answered all of them positively. The results of our TCRC shows the possibility to handle chaotic time series even with a need for memory in the framework of reservoir computing without the need of random initializations. Still our results show the benefit of reintroduction of randomness as $W^{\textrm{in}}$ in the TCRC-ELM for the prediction of chaotic time series.
	
	\paragraph{Limitations of the study}
	The utilized time series, i.e. approximated variations of the Mackey-Glass equations, are not representative for all problems. For example, the data do not suffer from noise or faulty measurements. We have limited ourselves to approximated time series to allow for interpretability of the results because of known characteristics of the time series. Furthermore, we did not use multivariate time series due to the rare occurrence of delayed differential equations in multiple dimensions in general and specifically we are not aware of a single one that is used as benchmark in previous RC studies. The training length $S_T$ was set to a fixed value for the sake of comparability.
	Several studies propose the use of a leakage rate as addition to the initial ESN and also demonstrate that the concept can yield substantially better performing networks for some datasets. We aim to compare to the fundamental version of the ESN as baseline and therefore decided to not utilize a leakage.
	Furthermore, recent studies also proposed alternative approaches to ESNs regarding the initialization of the random matrices or alternative topologies of these were also not used as baselines because of the limited comparability in literature and their partially specialised architecture for certain characteristics of time series. 
	For the Mackey-Glass-based datasets utilized for testing our approaches we have chosen an approximation with a step width of one. This leads to the trajectory itself becoming not-periodic for $\tau=15$, which can lead to the results varying, especially for this delay $\tau$, between approximations.
	We also remark that differences in the implementation allow different runtimes. An extensive comparison of methods of the software engineering as it would be necessary to show influences of such is the focus of currently conducted research.
	
	\section{Conclusion}\label{sec:concl}
	The prediction of time series is a challenging task even for modern neural networks, becoming even harder time series with long historic dependences and chaotic behavior. An architecture that has been demonstrated well suited for this kind of task is the mostly randomly initialized RC. In the last years, also TCN have been demonstrated as an alternative approach to handle history dependent time series, which in contrast need a more complex iterative learning mechanism. Inspired by the learning mechanism of RC and the history handling of TCN, we propose two novel network architectures. The architectures utilize a novel deterministic mapping of inputs into the network's state space, a method to create deep, multi-layered RC networks based on the deterministic mapping, and finally the additional use of a random mapping based on the size of the non-random network prior. The proposed methods improve the predictive accuracy compared to ESNs, as the most prominent RC architectures, by several orders of magnitude. We also observe a benefit of depth, i.e. multi-layered architectures, in case of the TCRC yielding results that outperform the ESN for all tested time series by up to $99.99\%$ for non-chaotic and $46.88\%$ for chaotic time series. If optimized in regard to its number of layers the TCRC outperforms even the GRU with the exceptions of $\tau\in\{30\}$. 
	The TCRC-ELM outperforms the ESN for non-chaotic time series by a similar relative improvement, for the chaotic ones up $85.45\%$. With the exception of $\tau=20$ the TCRC-ELM outperforms all other networks excepts the AEESN, including the GRU for the chaotic time series. 
	These results were achieved with an optimized ESN but without the optional usage of a leakage rate as additional hyper-parameter.
	
	We plan to study additional scalings and a linear blending between state spaces at different time steps for our proposed method. An extensive and expensive study with delayed ELMs, NG-RCs with large $\hat{\delta}$ and comparatively scaled states spaces for TCRC-ELMs is also seen as future work. Further, we plan to explore the possibilities of non-random mappings for the TCRC-ELM to extinguish the randomness and need for multiple runs, reducing the runtime substantially.

	\section*{Funding}
	We are funded by the Carl Zeiss Foundation's grants: P2018-02-001 %DeepTurb
	and P2017-01-005 %E4SM
	and by the German Federal Ministry of Education and Research (BMBF) grant: 02P22A040. 
	%	\section*{Data Availability}
	%	The authors agree to share the data and code upon reasonable request.
	
	%\section*{Declaration of Competing Interest}
	%The authors declare that they have no known competing financial interests or personal relationships that could have appeared to influence the work reported in this paper.

	\bibliographystyle{cas-model2-names}
	\bibliography{hyper.bib}

	\pagebreak
	\appendix
	\section{Prediction of Hyperchaotic Mackey-Glass Equation}
	The hyper-parameters of the ESN, GRU, AEESN, TCRC and TCRC-ELM were chosen as in their respective proposals \cite{ma2020deepr}, with testing a variational autoencoder as well as a Primary Component Analysis as substitute of an autoencoder. We reported the best combination of hyper-parameters, found as described in Sec. \ref{sec:eval} in Tab. \ref{tab:app:AEESN}. The reported value of $\tau=30$ marks the threshold for the step from chaotic to hyperchaotic behavior with a second positive Lyapunov-exponent. 
	\begin{table*}
		\centering
		\caption{MSE of best performing parametrizations for the tested networks for prediction of $286$ time steps for $\tau = 30$ of the Mackey-Glass equation. }
		
		\begin{tabular} {c|c|c|c|c|c}
			\toprule
			$\boldsymbol{\tau}$ &\textbf{ESN}  &\textbf{AEESN}   &\textbf{GRU}  &\textbf{TCRC} &\textbf{TCRC-ELM} 
			\\ 
			\midrule
			{$30$} &$1.15\cdot10^{1}$& \boldsymbol{$3.11\cdot10^{-2}$}  &$3.60\cdot10^{-1}$&$7.13\cdot10^{-2}$&$6.73\cdot10^{-2}$\\
			\bottomrule
		\end{tabular}
		\label{tab:app:AEESN}
	\end{table*}

	\section{Visual Comparison of predictions}
	We show an exemplary comparison, limited to one run of each predictor, for the best prediction, measured by MSE.
	\begin{figure*}[htb]
		\centering
		%\caption*{Non-chaotic variation of the Mackey-Glass time series}
		\begin{subfigure}{.32\textwidth}
			\includegraphics[clip,width=\textwidth]{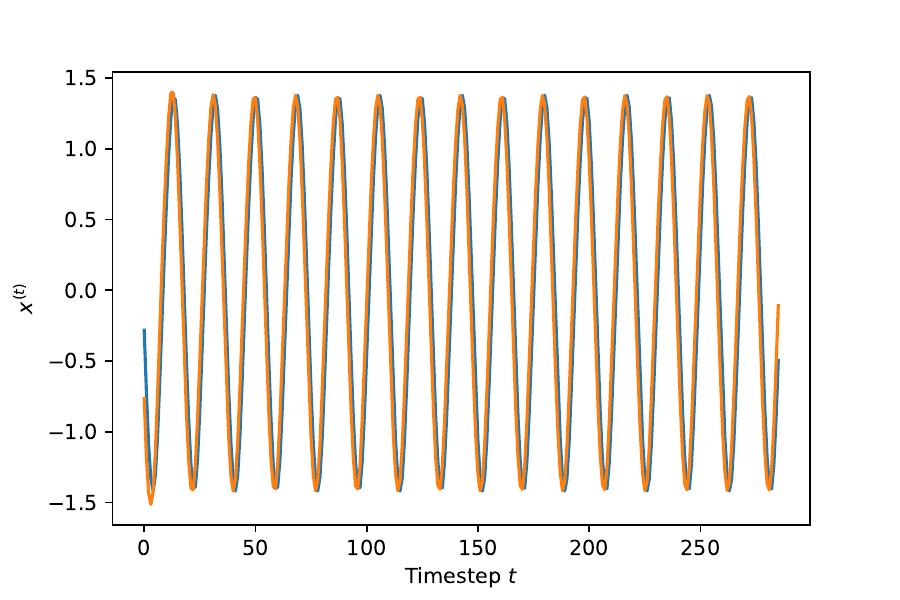}
			\caption{$\tau=5$}
		\end{subfigure}
		\begin{subfigure}{.32\textwidth}
			\includegraphics[clip,width=\textwidth]{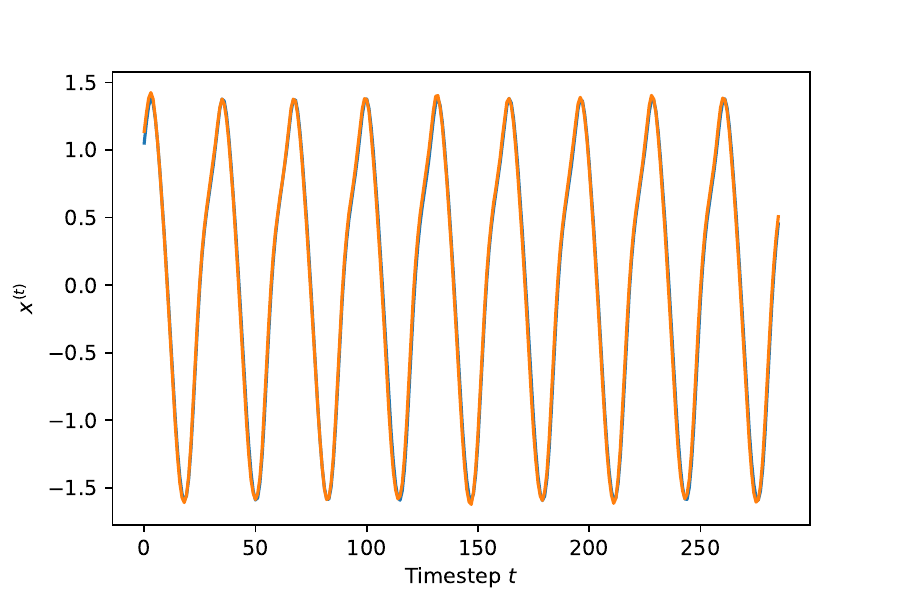}
			\caption{$\tau=10$}
		\end{subfigure}
		\begin{subfigure}{.32\textwidth}
			\includegraphics[clip,width=\textwidth]{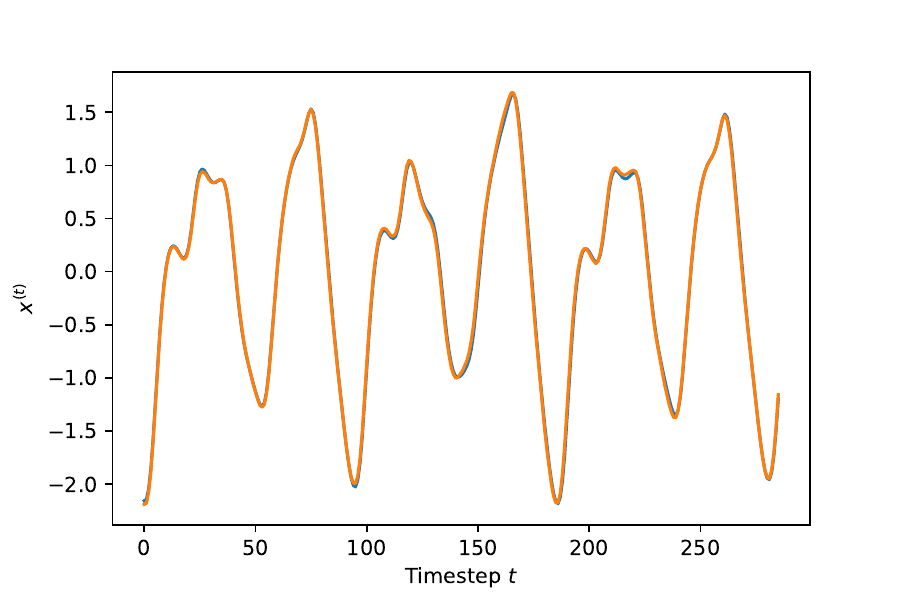}
			\caption{$\tau=15$}
		\end{subfigure}
		%\caption*{Chaotic variation of the Mackey-Glass time series}
		\begin{subfigure}{.32\textwidth}
			\includegraphics[clip,width=\textwidth]{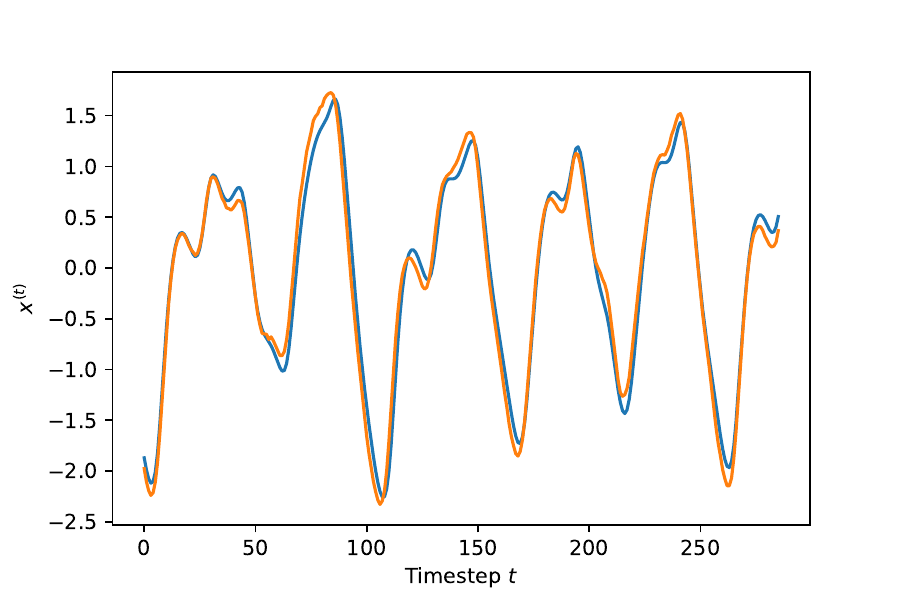}
			\caption{$\tau=17$}
		\end{subfigure}
		\begin{subfigure}{.32\textwidth}
			\includegraphics[clip,width=\textwidth]{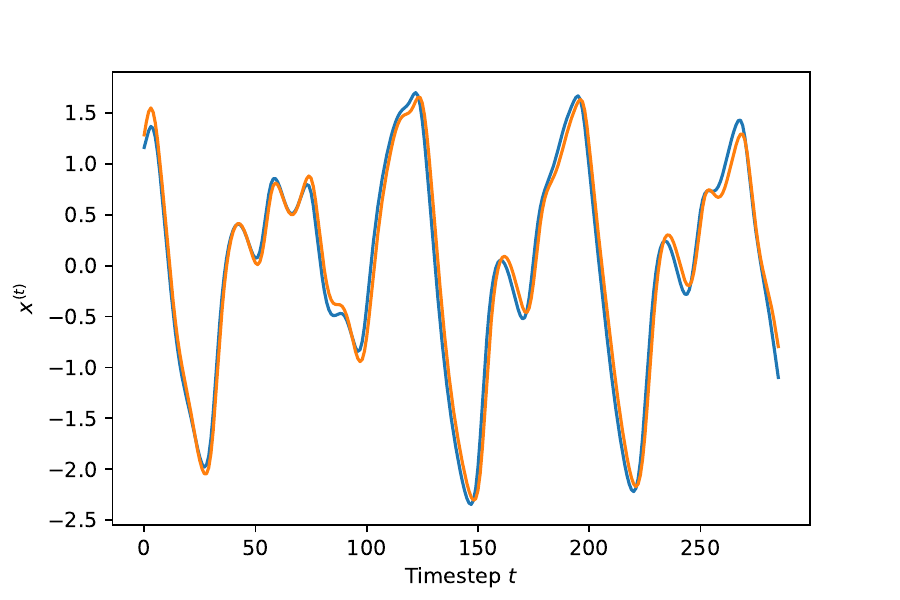}
			\caption{$\tau=20$}
		\end{subfigure}
		\begin{subfigure}{.32\textwidth}
			\includegraphics[clip,width=\textwidth]{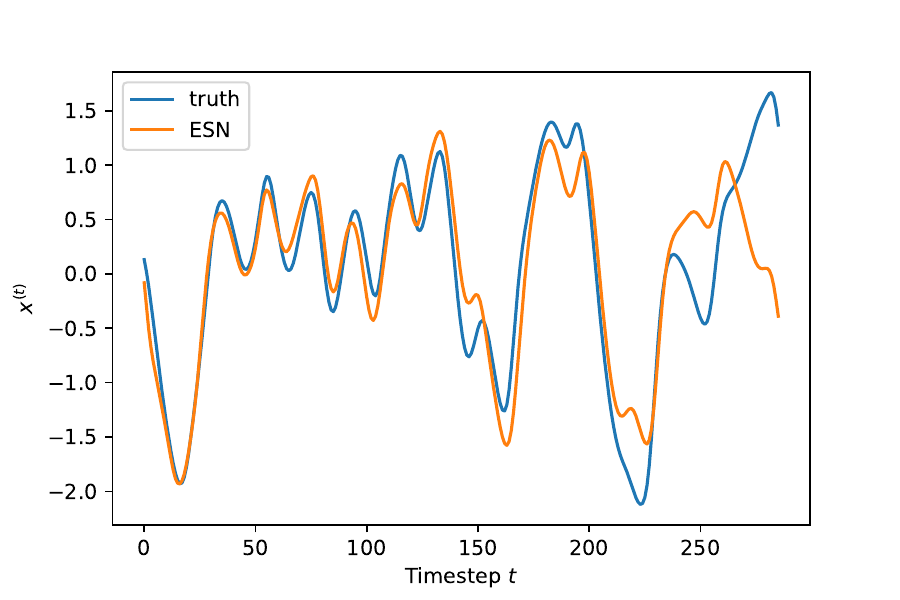}
			\caption{$\tau=25$}
		\end{subfigure}
		\caption{Exemplary visualization of the predicted Mackey-Glass equation for $286$ time steps and different values of $\tau$ with ESN. }% \hl{influencing the chaoticity of the time series -- korrekt???}.}
		\label{fig:app:pred_plots_ESN}
	\end{figure*}
	\begin{figure*}[htb]
		\centering
		%\caption*{Non-chaotic variation of the Mackey-Glass time series}
		\begin{subfigure}{.32\textwidth}
			\includegraphics[clip,width=\textwidth]{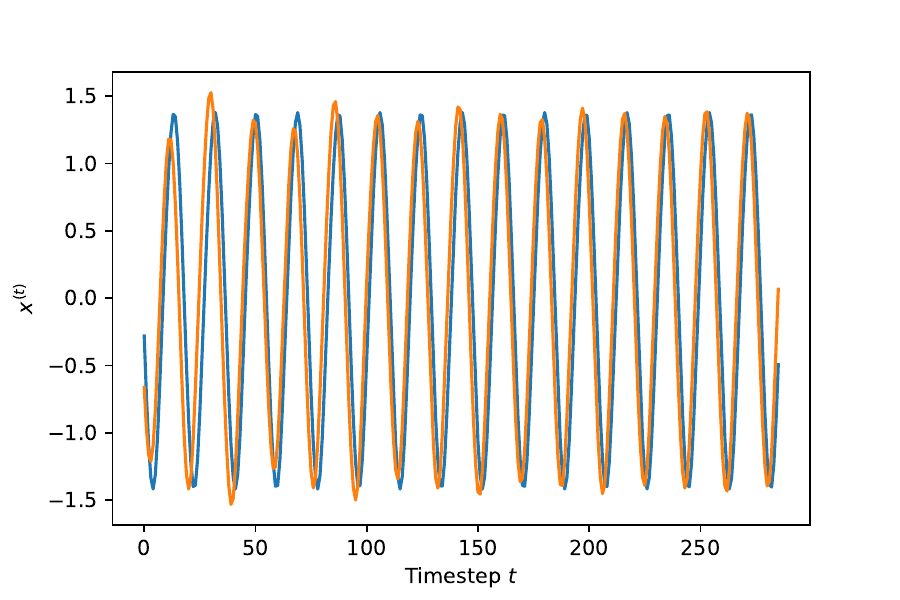}
			\caption{$\tau=5$}
		\end{subfigure}
		\begin{subfigure}{.32\textwidth}
			\includegraphics[clip,width=\textwidth]{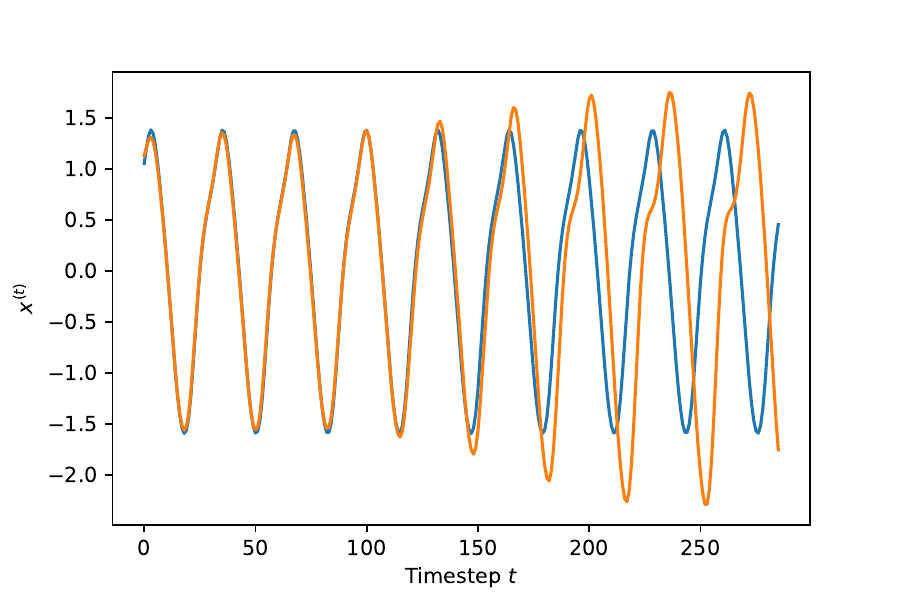}
			\caption{$\tau=10$}
		\end{subfigure}
		\begin{subfigure}{.32\textwidth}
			\includegraphics[clip,width=\textwidth]{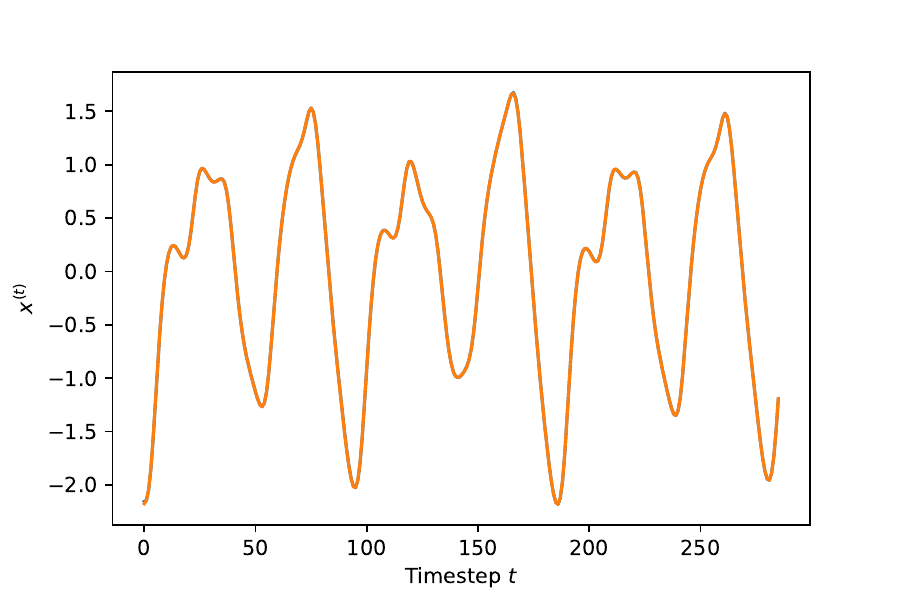}
			\caption{$\tau=15$}
		\end{subfigure}
		%\caption*{Chaotic variation of the Mackey-Glass time series}
		\begin{subfigure}{.32\textwidth}
			\includegraphics[clip,width=\textwidth]{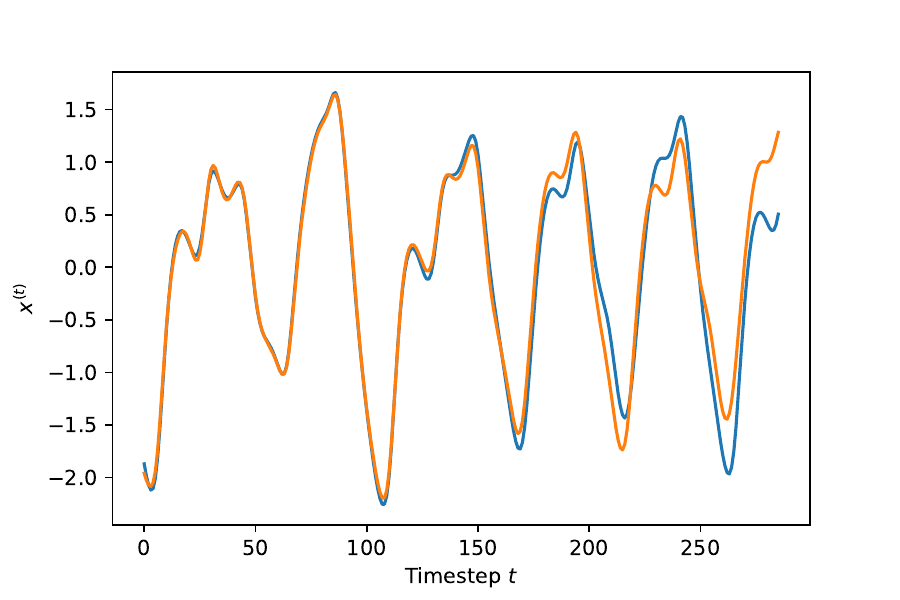}
			\caption{$\tau=17$}
		\end{subfigure}
		\begin{subfigure}{.32\textwidth}
			\includegraphics[clip,width=\textwidth]{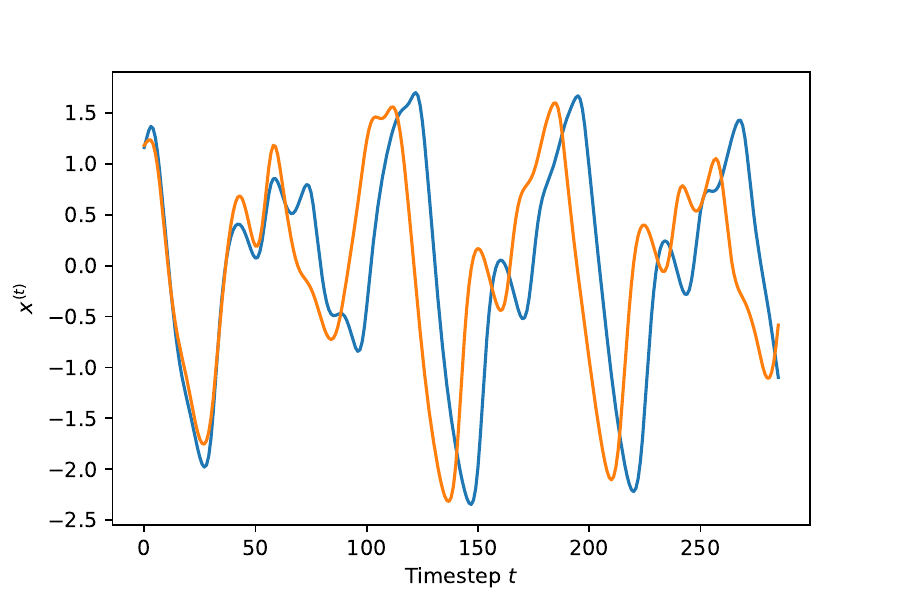}
			\caption{$\tau=20$}
		\end{subfigure}
		\begin{subfigure}{.32\textwidth}
			\includegraphics[clip,width=\textwidth]{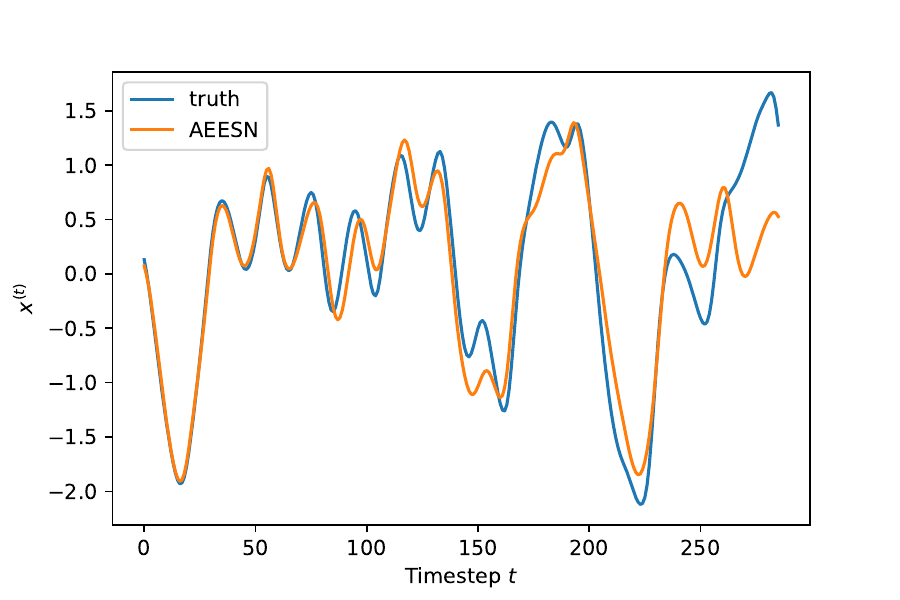}
			\caption{$\tau=25$}
		\end{subfigure}
		\caption{Exemplary visualization of the predicted Mackey-Glass equation for $286$ time steps and different values of $\tau$ with AEESN. }% \hl{influencing the chaoticity of the time series -- korrekt???}.}
		\label{fig:app:pred_plots_AEESN}
	\end{figure*}
	\begin{figure*}[htb]
		\centering
		%\caption*{Non-chaotic variation of the Mackey-Glass time series}
		\begin{subfigure}{.32\textwidth}
			\includegraphics[clip,width=\textwidth]{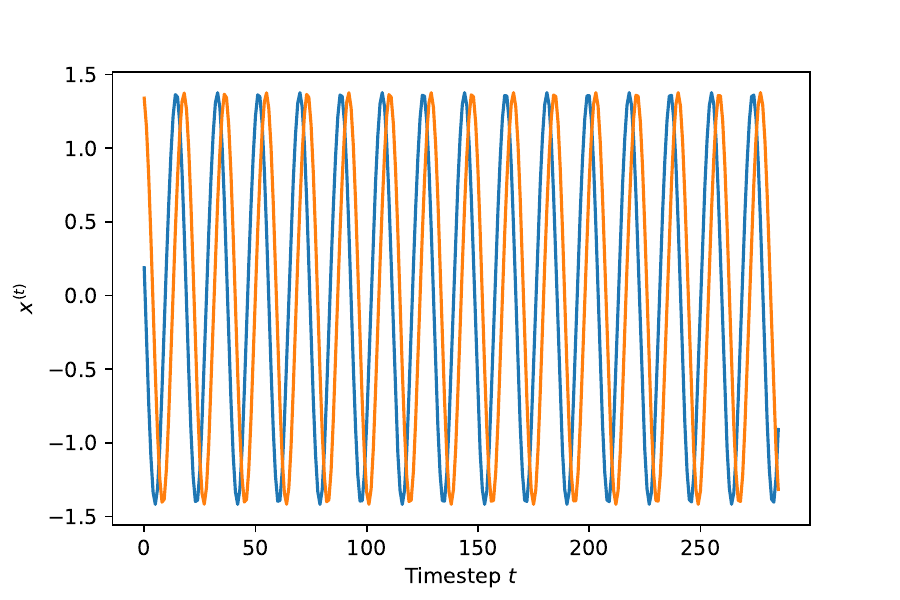}
			\caption{$\tau=5$}
		\end{subfigure}
		\begin{subfigure}{.32\textwidth}
			\includegraphics[clip,width=\textwidth]{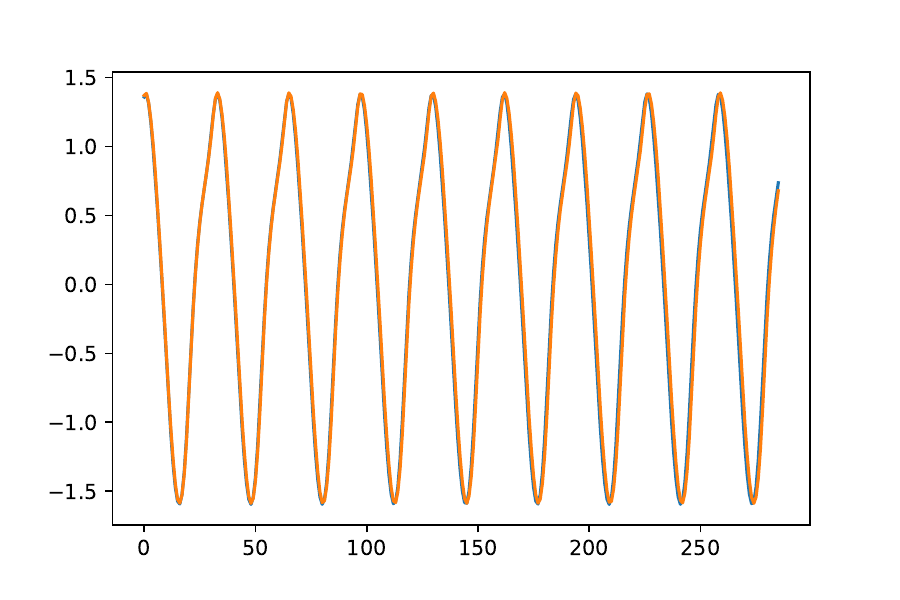}
			\caption{$\tau=10$}
		\end{subfigure}
		\begin{subfigure}{.32\textwidth}
			\includegraphics[clip,width=\textwidth]{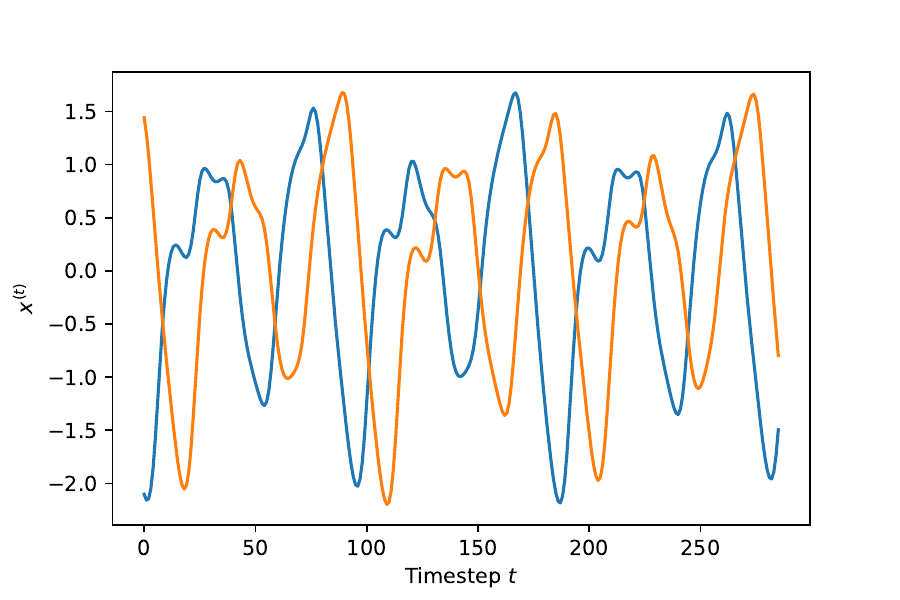}
			\caption{$\tau=15$}
		\end{subfigure}
		%\caption*{Chaotic variation of the Mackey-Glass time series}
		\begin{subfigure}{.32\textwidth}
			\includegraphics[clip,width=\textwidth]{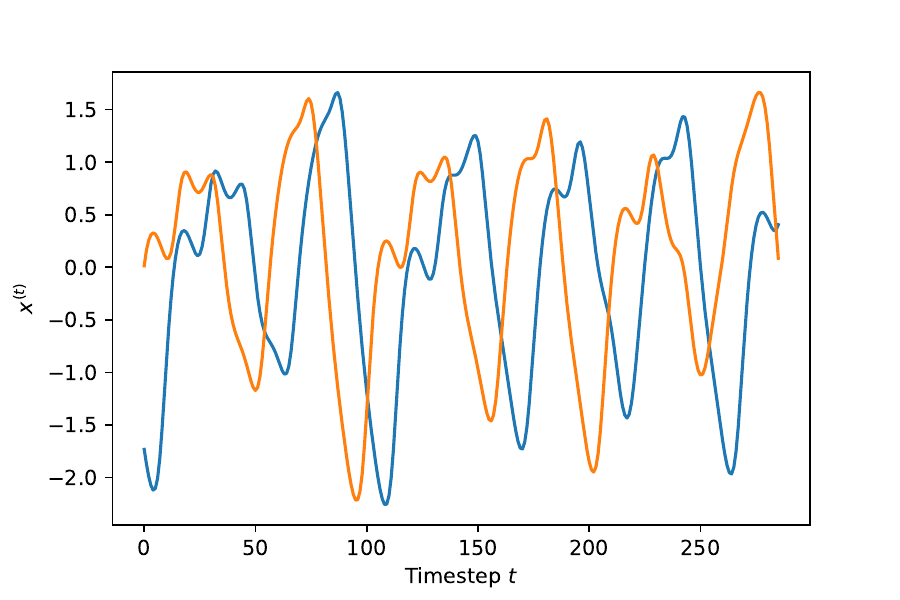}
			\caption{$\tau=17$}
		\end{subfigure}
		\begin{subfigure}{.32\textwidth}
			\includegraphics[clip,width=\textwidth]{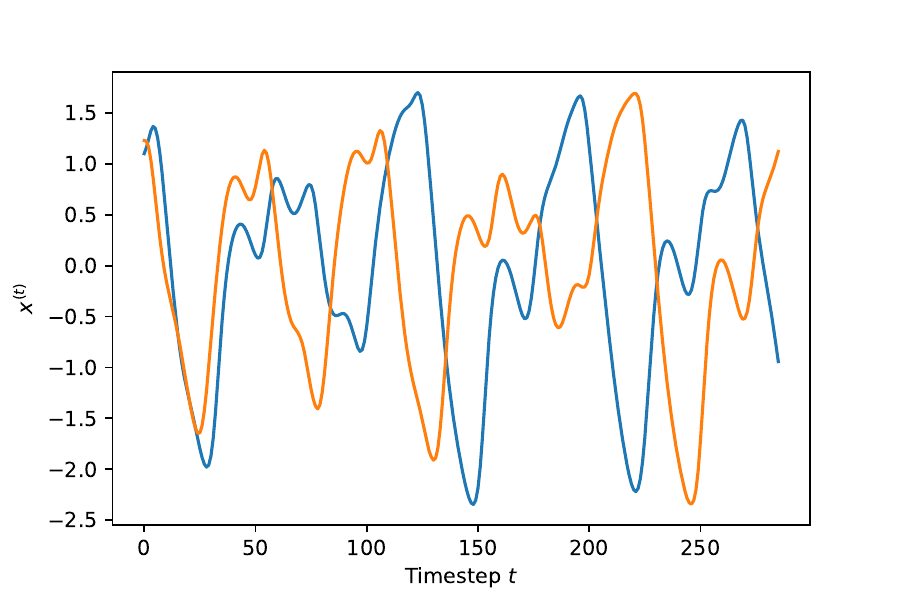}
			\caption{$\tau=20$}
		\end{subfigure}
		\begin{subfigure}{.32\textwidth}
			\includegraphics[clip,width=\textwidth]{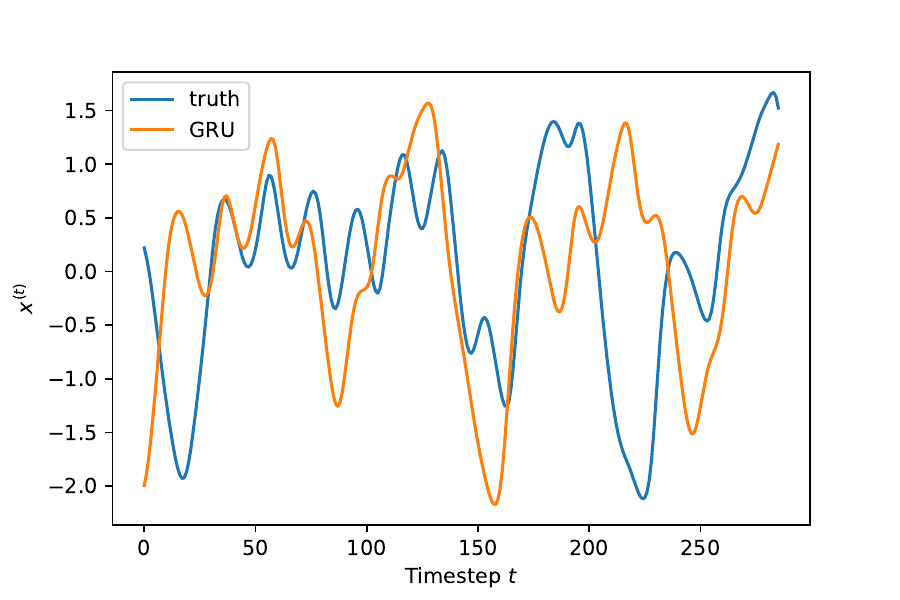}
			\caption{$\tau=25$}
		\end{subfigure}
		\caption{Exemplary visualization of the predicted Mackey-Glass equation for $286$ time steps and different values of $\tau$ with GRU. }% \hl{influencing the chaoticity of the time series -- korrekt???}.}
		\label{fig:app:pred_plots_GRU}
	\end{figure*}
	\begin{figure*}[htb]
		\centering
		%\caption*{Non-chaotic variation of the Mackey-Glass time series}
		\begin{subfigure}{.32\textwidth}
			\includegraphics[clip,width=\textwidth]{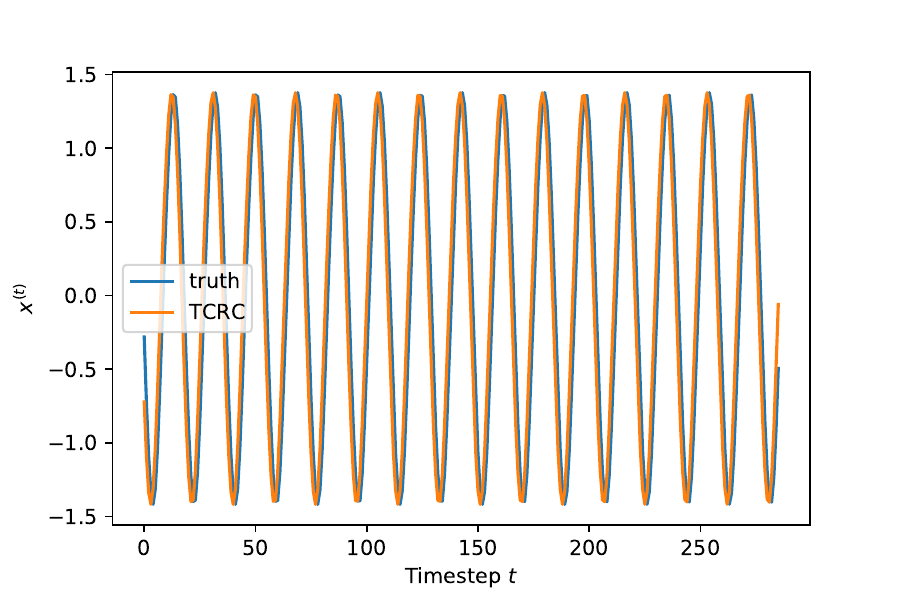}
			\caption{$\tau=5$}
		\end{subfigure}
		\begin{subfigure}{.32\textwidth}
			\includegraphics[clip,width=\textwidth]{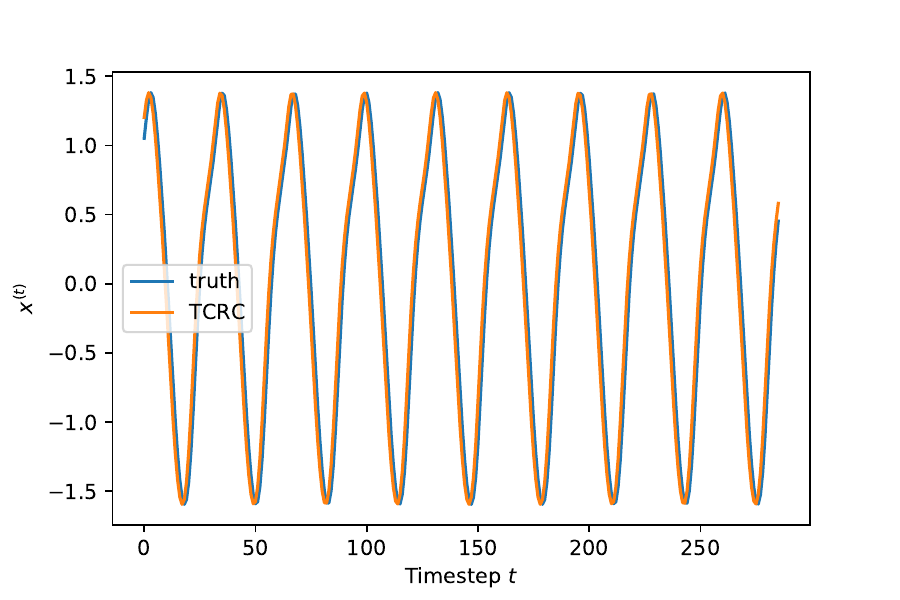}
			\caption{$\tau=10$}
		\end{subfigure}
		\begin{subfigure}{.32\textwidth}
			\includegraphics[clip,width=\textwidth]{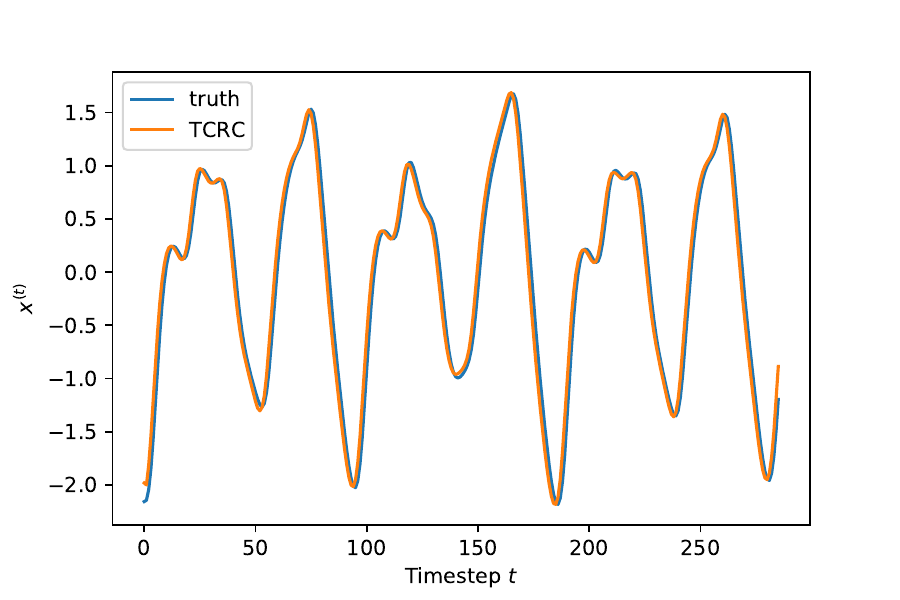}
			\caption{$\tau=15$}
		\end{subfigure}
		%\caption*{Chaotic variation of the Mackey-Glass time series}
		\begin{subfigure}{.32\textwidth}
			\includegraphics[clip,width=\textwidth]{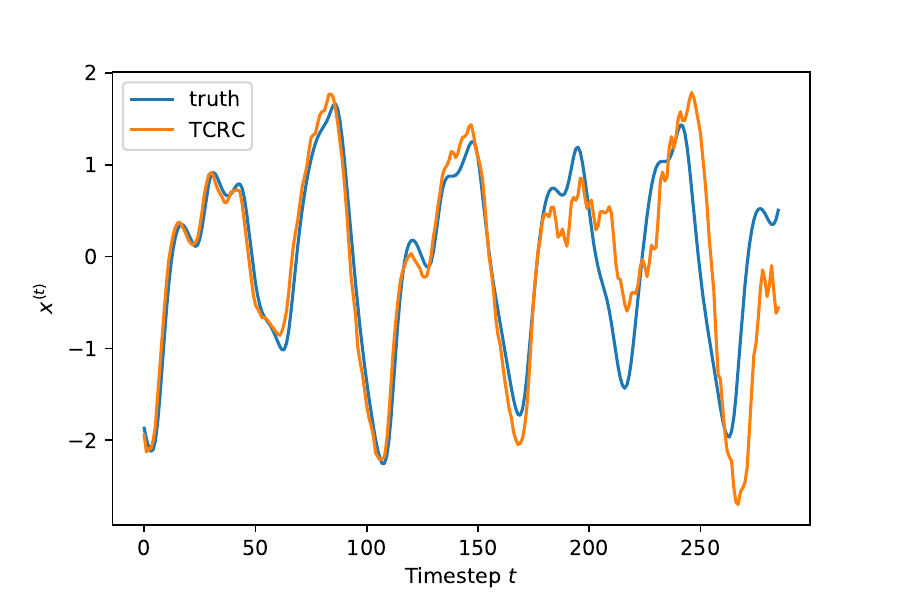}
			\caption{$\tau=17$}
		\end{subfigure}
		\begin{subfigure}{.32\textwidth}
			\includegraphics[clip,width=\textwidth]{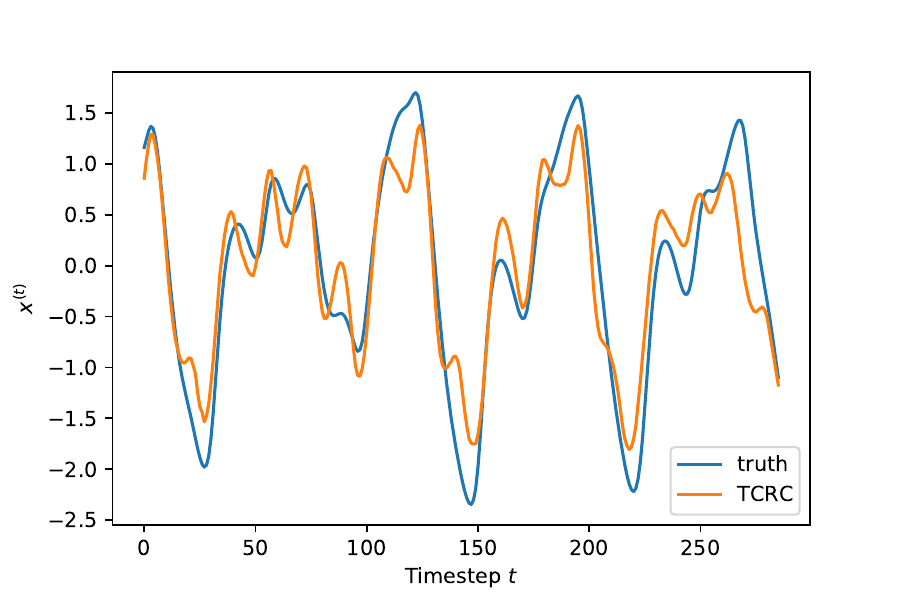}
			\caption{$\tau=20$}
		\end{subfigure}
		\begin{subfigure}{.32\textwidth}
			\includegraphics[clip,width=\textwidth]{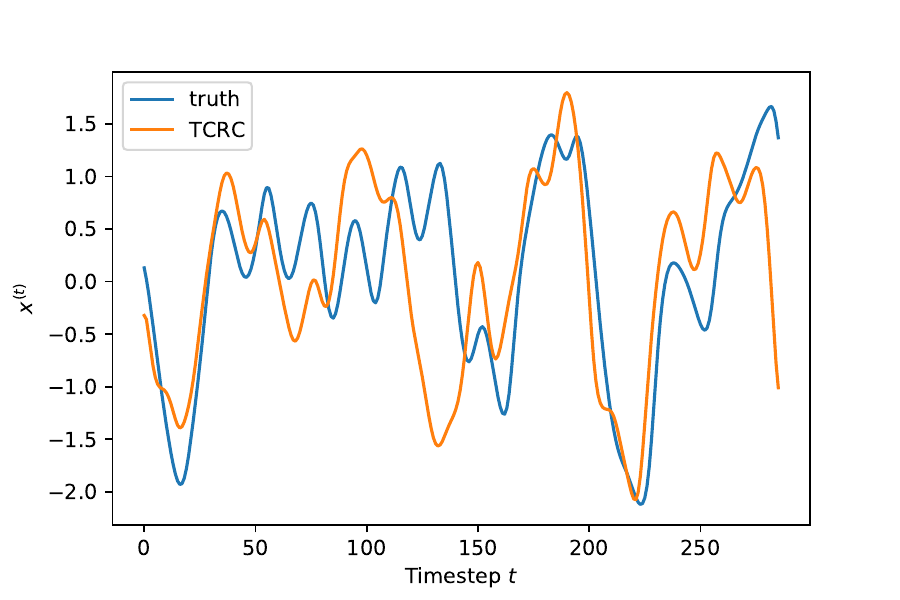}
			\caption{$\tau=25$}
		\end{subfigure}
		\caption{Exemplary visualization of the predicted Mackey-Glass equation for $286$ time steps and different values of $\tau$ with TCRC. }% \hl{influencing the chaoticity of the time series -- korrekt???}.}
		\label{fig:app:pred_plots_TCRC}
	\end{figure*}
	\begin{figure*}[htb]
		\centering
		%\caption*{Non-chaotic variation of the Mackey-Glass time series}
		\begin{subfigure}{.32\textwidth}
			\includegraphics[clip,width=\textwidth]{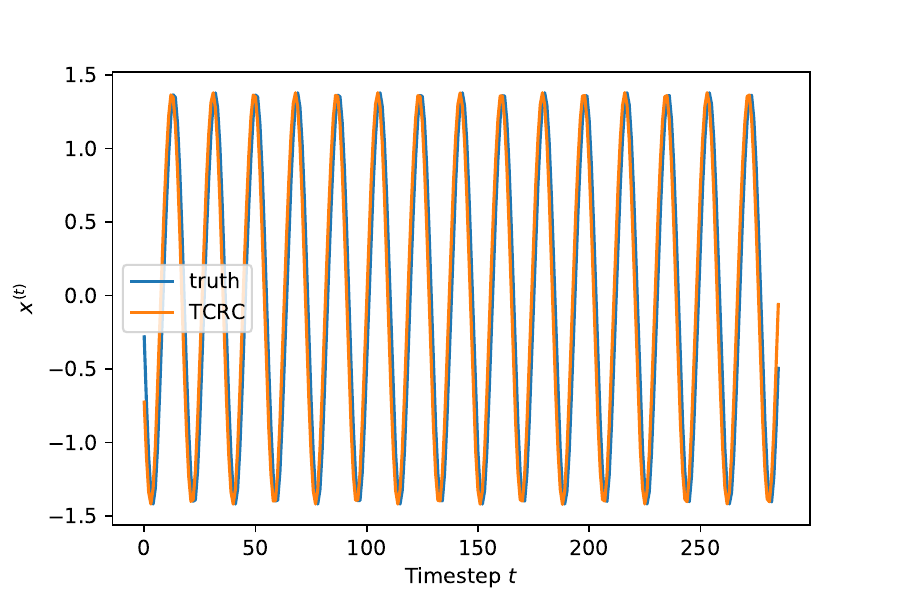}
			\caption{$\tau=5$}
		\end{subfigure}
		\begin{subfigure}{.32\textwidth}
			\includegraphics[clip,width=\textwidth]{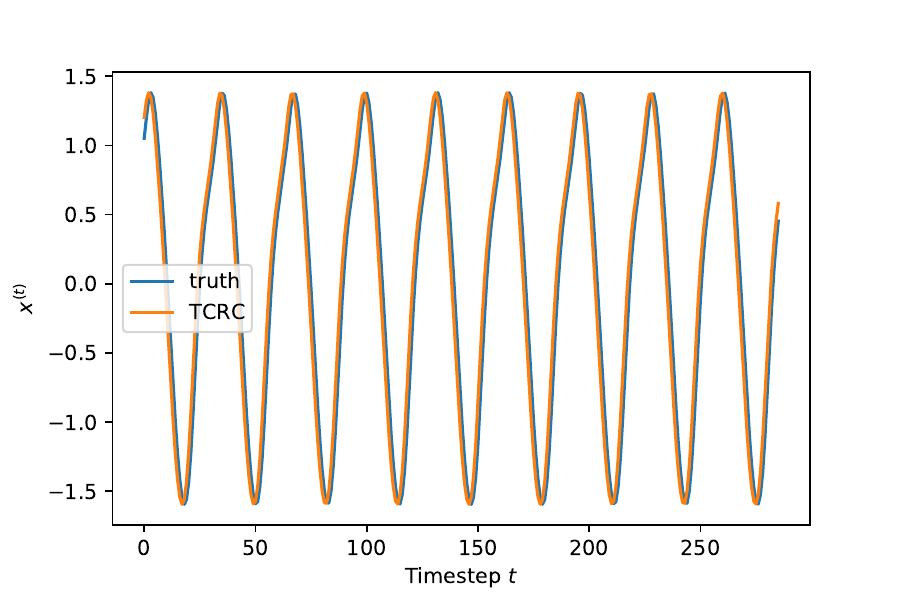}
			\caption{$\tau=10$}
		\end{subfigure}
		\begin{subfigure}{.32\textwidth}
			\includegraphics[clip,width=\textwidth]{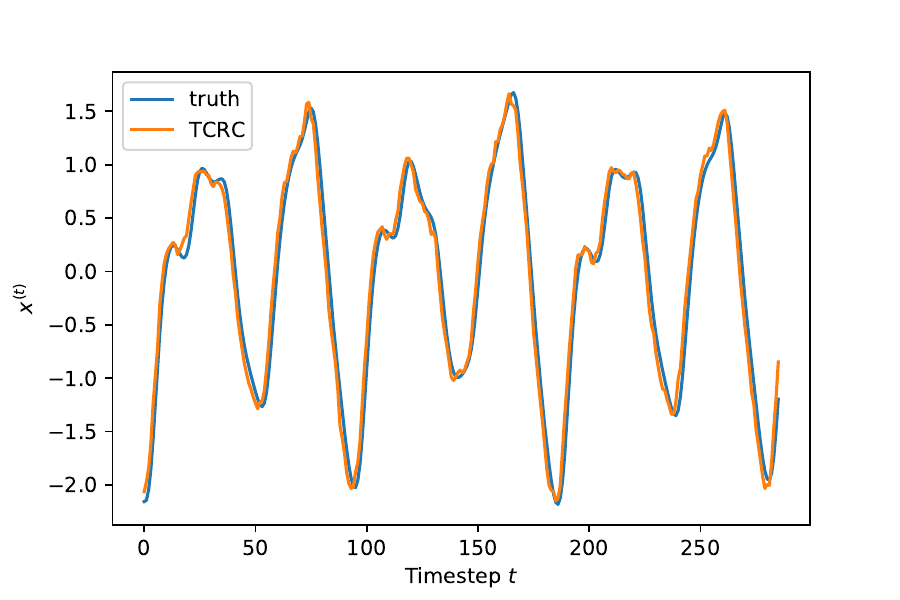}
			\caption{$\tau=15$}
		\end{subfigure}
		%\caption*{Chaotic variation of the Mackey-Glass time series}
		\begin{subfigure}{.32\textwidth}
			\includegraphics[clip,width=\textwidth]{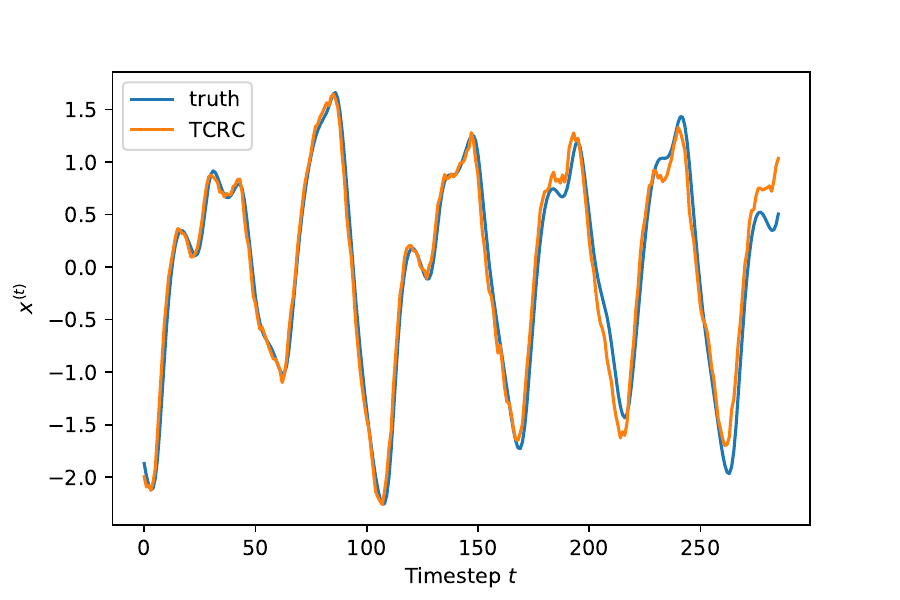}
			\caption{$\tau=17$}
		\end{subfigure}
		\begin{subfigure}{.32\textwidth}
			\includegraphics[clip,width=\textwidth]{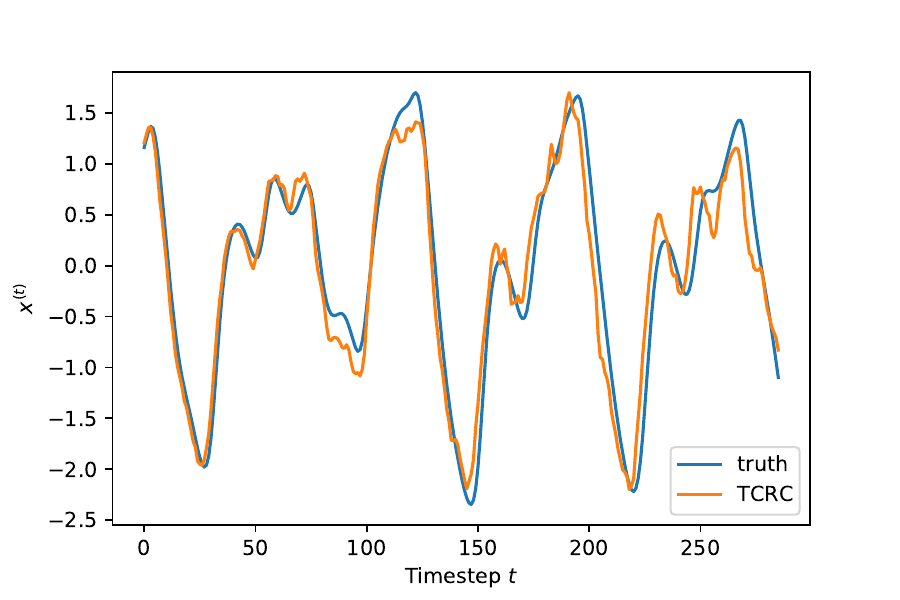}
			\caption{$\tau=20$}
		\end{subfigure}
		\begin{subfigure}{.32\textwidth}
			\includegraphics[clip,width=\textwidth]{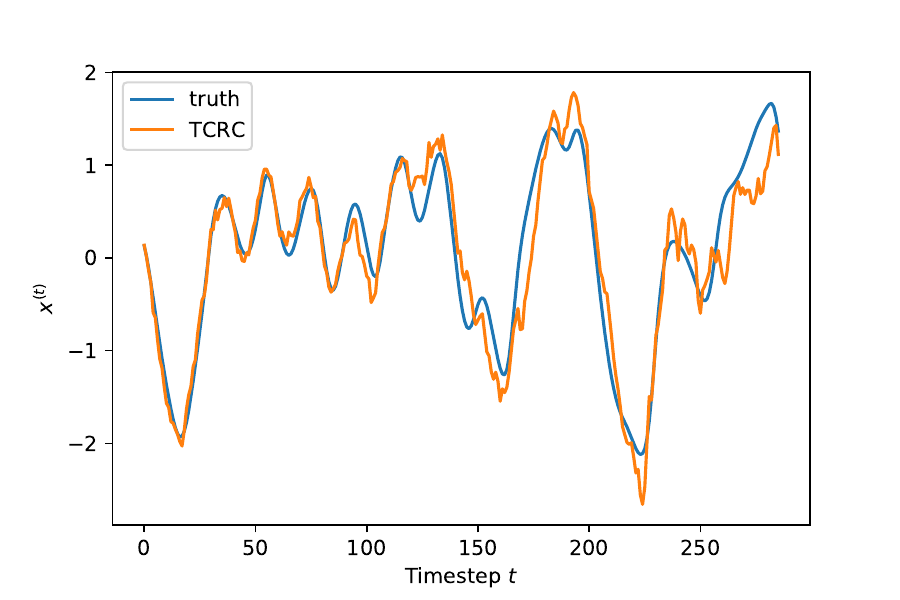}
			\caption{$\tau=25$}
		\end{subfigure}
		\caption{Exemplary visualization of the predicted Mackey-Glass equation for $286$ time steps and different values of $\tau$ with TCRC-ELM. }% \hl{influencing the chaoticity of the time series -- korrekt???}.}
		\label{fig:app:pred_plots_TCRCELM}
	\end{figure*}
	\section{Number of Valid Time Steps}
	\begin{table*}
		\centering
		\caption{Number of valid time steps of best performing Basic ESN, AEESN, GRU and TCRC for different $\tau$ of the Mackey-Glass equation, with valid being predictions within one standard deviation difference to the ground truth. We present the best tested $\rho\in[0,1.5]$ and $N^{\mathrm{r}}\in\{1000,2000,3000\}$ as well as best tested input lengths $\hat{\delta}$ for the GRU and TCRC.}
		\begin{adjustbox}{width=\textwidth}
			\begin{tabular} {c|c|c|c|c|c}
				\toprule
				$\boldsymbol{\tau}$ &\multicolumn{1}{>{\centering\arraybackslash}m{.192\textwidth}|}{\textbf{ESN} } &\multicolumn{1}{>{\centering\arraybackslash}m{.192\textwidth}|}{\textbf{AEESN} }  &\multicolumn{1}{>{\centering\arraybackslash}m{.192\textwidth}|}{\textbf{GRU} } &\multicolumn{1}{>{\centering\arraybackslash}m{.192\textwidth}|}{\textbf{TCRC} }&\multicolumn{1}{>{\centering\arraybackslash}m{.192\textwidth}}{\textbf{TCRC-ELM} }
				\\ 
				
				\midrule
				{$5$}& {$286$}&$16$&$286$&$286$ &$286$\\
				\midrule
				{$10$} & {$286$} &$286$&$286$& $286$ &$286$\\
				\midrule
				{$15$} & {$286$} &$286$&$0$&$286$  &$286$\\
				\midrule
				{$17$} & {$286$} &$286$&$115$&$209$ &$286$\\
				\midrule
				{$20$} & $286$ &$286$&$30$& $267$  &$286$\\
				\midrule
				{$25$} & {$273$}  &$286$&$286$&$100$ &$170$\\
				\bottomrule
			\end{tabular}
		\end{adjustbox}
		\label{tab:app:valsteps}
	\end{table*}

\end{document}